\documentclass[lettersize,journal]{IEEEtran}
\usepackage{amsmath,amsfonts,amssymb}
\usepackage{array}
\usepackage[caption=false,font=normalsize,labelfont=sf,textfont=sf]{subfig}
\usepackage{textcomp}
\usepackage{stfloats}
\usepackage{url}
\usepackage{verbatim}
\usepackage{graphicx}
\usepackage{cite}
\usepackage{pifont}
\usepackage{authblk}
\usepackage{enumitem}

\usepackage{amsthm}

\usepackage{booktabs}
\usepackage{multirow}
\usepackage{lipsum}
\usepackage{hyperref}
\usepackage[bottom]{footmisc}
\usepackage{url}
\usepackage{breakurl}

\usepackage[edges]{forest}
\definecolor{hidden-draw}{RGB}{20,68,106}
\definecolor{hidden-pink}{RGB}{255,245,247}
\usepackage[framemethod=tikz]{mdframed}
\usepackage{subcaption}
\usepackage{soul}
\usepackage{times}
\usepackage{tabularx}
\usepackage{float}
\usepackage{xcolor} 
\usepackage{latexsym}
\usepackage{bbm}
\usepackage{amsfonts}
\usepackage{dsfont}
\usepackage{graphicx}
\usepackage{caption}
\usepackage{subcaption}
\usepackage{algpseudocode}
\usepackage{colortbl}
\definecolor{lightgray}{gray}{0.9}
\definecolor{lightgreen}{rgb}{0.9, 1, 0.9}
\usepackage{microtype}
\usepackage{inconsolata}

\usepackage{amsmath,amsfonts,bm}

\def\eqref#1{equation~\ref{#1}}

\def\1{\bm{1}}

\DeclareMathAlphabet{\mathsfit}{\encodingdefault}{\sfdefault}{m}{sl}
\SetMathAlphabet{\mathsfit}{bold}{\encodingdefault}{\sfdefault}{bx}{n}

\usepackage{tikz}
\usetikzlibrary{fadings}
\usetikzlibrary{mindmap,trees}
\usetikzlibrary{arrows,automata,shapes,positioning,shadows,trees}
\usetikzlibrary{shapes,snakes,shadows}
\usetikzlibrary{shapes.arrows}
\usetikzlibrary{calc,shapes, positioning}
\usetikzlibrary{decorations.text}
\usetikzlibrary{matrix,chains,positioning,decorations.pathreplacing,arrows}
\usetikzlibrary{bayesnet}
\usetikzlibrary{shadows.blur}
\usetikzlibrary{shapes.geometric}

\usepackage{pgfplots}
\usepackage{pgfplotstable}
\usepackage{pgfmath,pgffor}
\usepgfplotslibrary{colorbrewer}
\pgfplotsset{compat = 1.14, cycle list/Set1-8}
\usetikzlibrary{pgfplots.statistics, pgfplots.colorbrewer}
\pgfplotsset{compat=1.8}

\tikzstyle{edge}=[-latex',draw=black!90,shorten <=1pt,shorten >=1pt]
\tikzstyle{redge}=[latex'-,draw=black!90,shorten <=1pt,shorten >=1pt]
\tikzstyle{dedge}=[latex'-latex',draw=black!90,shorten <=1pt,shorten >=1pt]

\tikzstyle{block}=[draw, text width=5em,align=center,shape=rectangle, rounded corners, , align=center]
\tikzstyle{nobox}=[align=center]
\definecolor{emb}{RGB}{209,228,252}
\definecolor{hidden-blue}{RGB}{194,232,247}
\definecolor{hidden-orange}{RGB}{224,224,224}
\definecolor{hidden-yellow}{RGB}{242,244,193}
\definecolor{output-purple}{RGB}{219,203,231}
\definecolor{output-green}{RGB}{204,231,207}
\definecolor{hiddendraw}{RGB}{10,128,122}
\definecolor{myred2}{HTML}{F875AA}
\definecolor{mypurple2}{HTML}{D2E0FB}

\definecolor{myred}{HTML}{F8F6F4}
\definecolor{mypurple}{HTML}{FFDFDF}
\definecolor{myyellow}{HTML}{FFF6F6}
\definecolor{mygreen}{HTML}{D2E0FB}
\usepackage{todonotes}

\newtheorem{definition}{Definition}[section]

\usepackage{todonotes}

\begin{document}

\title{Graph-based Agent Memory: \\Taxonomy,
Techniques, and Applications}

\author{
\IEEEauthorblockN{
\textbf{Chang~Yang$^\dagger$, Chuang~Zhou$^\dagger$, Yilin~Xiao$^\dagger$, Su~Dong, Luyao~Zhuang, Yujing~Zhang, Zhu~Wang} \\
\vspace{-2ex}
\textbf{Zijin~Hong, Zheng~Yuan, Zhishang~Xiang, Shengyuan~Chen$^\ddagger$, Huachi~Zhou$^\ddagger$, Qinggang~Zhang$^\ddagger$} \\
\textbf{Ninghao~Liu, Jinsong~Su, Xinrun~Wang, Yi~Chang, Xiao~Huang}
\thanks{$\dagger$Equal contribution.}
\thanks{$\ddagger$Corresponding authors: Qinggang Zhang, Huachi Zhou, Shengyuan Chen.}
\thanks{Chang Yang, Chuang Zhou, Yilin Xiao, Su Dong, Luyao Zhuang, Yujing Zhang, Zhu Wang, Zijin Hong, Zheng Yuan, Shengyuan Chen, Huachi Zhou, Qinggang Zhang, Ninghao Liu, and Xiao Huang are with the The Hong Kong Polytechnic University, Hong Kong SAR, China (e-mail: \{chang.yang, chuang-qqzj.zhou, yilin.xiao, su.dong, luyao.zhuang, yu-jing.zhang, juliazhu.wang,  zijin.hong, yzheng.yuan, huachi.zhou\}@connect.polyu.hk, \{sheng-yuan.chen, qinggang.zhang, ninghao-prof.liu, xiao.huang\}@polyu.edu.hk).}
\thanks{Jinsong Su and Zhishang Xiang are with the School of Information, Xiamen University, China (e-mail: xiangzhishang@stu.xmu.edu.cn, jssu@xmu.edu.cn).}
\thanks{Xinrun Wang is with the School of Computing and Information Systems, Singapore Management University, Singapore (e-mail: xrwang@smu.edu.sg).}
\thanks{Yi Chang is with the School of Artificial Intelligence, Jilin University, Changchun, China (e-mail: yichang@jlu.edu.cn).}
}}


\maketitle


\begin{abstract}

Memory emerges as the core module in the Large Language Model (LLM)-based agents for long-horizon complex tasks (e.g., multi-turn dialogue, game playing, scientific discovery), where memory can enable knowledge accumulation, iterative reasoning and self-evolution. Among diverse paradigms, graph stands out as a powerful structure for agent memory due to the intrinsic capabilities to model relational dependencies, organize hierarchical information, and support efficient retrieval. This survey presents a comprehensive review of agent memory from the graph-based perspective. First, we introduce a taxonomy of agent memory, including short-term vs. long-term memory, knowledge vs. experience memory, non-structural vs. structural memory, with an implementation view of graph-based memory. Second, according to the life cycle of agent memory, we systematically analyze the key techniques in graph-based agent memory, covering memory extraction for transforming the data into the contents, storage for organizing the data efficiently, retrieval for retrieving the relevant contents from memory to support reasoning, and evolution for updating the contents in the memory. 
Third, we summarize the open-sourced libraries and benchmarks that support the development and evaluation of self-evolving agent memory. We also explore diverse application scenarios.
Finally, we identify critical challenges and future research directions.
This survey aims to offer actionable insights to advance the development of more efficient and reliable graph-based agent memory systems. All the related resources, including research papers, open-source data, and projects, are collected for the community in \textcolor{blue}{\url{https://github.com/DEEP-PolyU/Awesome-GraphMemory}}.

\end{abstract}


\ifCLASSOPTIONcompsoc
\IEEEraisesectionheading{\section{Introduction}\label{sec:introduction}}
\else

\begin{figure*}[ht]
	\centering
    \includegraphics[width=\linewidth]{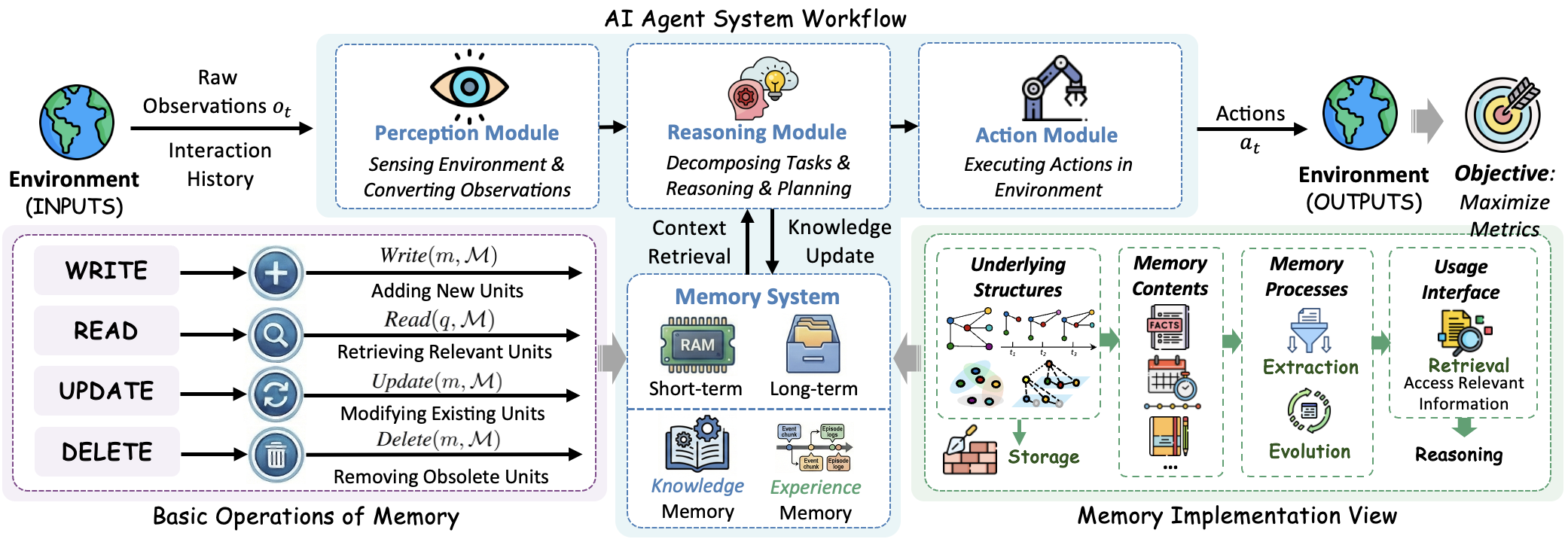}
    \caption{A diagram illustrating the workflow of an AI agent system, and the detailed implementations of the memory system.} 
    \label{fig:ai_agent}
\end{figure*}

\section{Introduction}

The past few years witnessed the rapid development of Large Language Model (LLM)-based agents, which have demonstrated remarkable success in complex, long-horizon tasks across diverse domains, from software engineering~\cite{yang2024swe} and mathematical reasoning~\cite{lin2025goedel} to multi-agent tasks~\cite{meta2022human} and open-world exploration~\cite{wang2024voyager}. 
The inherent language understanding, generation, and inference capabilities of LLMs enable LLM-based agents to autonomously perceive environments and make decisions, which reduces manual intervention and reshapes the paradigm of intelligent systems~\cite{su2024language}.

Despite notable advancements, LLM-based agents are still constrained by the intrinsic limitations of LLMs.
(i) Knowledge cutoff: LLMs are trained on static datasets with fixed time boundaries, resulting in knowledge cutoff issues that prevent them from incorporating real-time information (e.g., current financial data) or domain-specific knowledge beyond their pre-training corpora. This limitation undermines their ability to adapt to dynamic environments and open-ended scenarios.
(ii) Tool incompetence: Although tool use represents a core capability of LLM-based agents~\cite{wang2024what,qin2024toolllm}, existing LLMs demonstrate limited capacity for efficiently learning and applying novel tools, which substantially constrains the agent performance.
(iii) Performance saturation:
LLM-based agents exhibit persistent failures in iterative, long-horizon tasks due to their inability to accumulate task-specific insights and leverage historical experiences for refining decision strategies across extended interactions. Consequently, agents may repeatedly commit similar errors without demonstrating learning behavior to correct the errors for successful task completion.


To address these challenges, memory~\cite{sumers2023cognitive} has emerged as a critical component for advancing LLM agents towards four key objectives: i) \textbf{Personalization and Specification.}~\cite{li2024personal}: Memory enables agents to capture user preferences, interaction histories, and task-specific contexts for tailored responses, such as remembering workflow habits in software engineering or communication styles in conversational scenarios. Memory bridges general knowledge with specific context, storing both universal facts and particular histories to ground responses in personalized, context-aware information~\cite{yang2025efficient}. ii) \textbf{Long-term Reasoning beyond Context Window.} While LLMs operate within finite context windows with static parametric knowledge, memory systems provide unbounded external storage that enables continuous learning and adaptation. Memory allows agents to retain information across extended temporal horizons, accumulate post-deployment experiences including successes and failures, and dynamically refine strategies without model retraining. iii) \textbf{Self-improving}\cite{shang2025agentsquare}: By accumulating experiential knowledge, reasoning patterns, and feedback, agent memory supports iterative enhancement of adaptability and performance, thus enabling the self-improving of LLM-based Agents on the tasks without parameter updating. iv) \textbf{Hallucination mitigation}~\cite{ye2023cognitive}: Grounding outputs in structured, verifiable memory content reduces reliance on potentially unreliable parametric knowledge. In essence, memory transforms stateless reactive models into stateful adaptive entities capable of relationship building, trajectory-based learning, and increasingly sophisticated personalized behavior.

Traditional implementations of agent memory primarily adopt linear, unstructured, or simple key-value storage paradigms, such as fixed-length token sequences, vector databases, and log-based buffers~\cite{yu2025memagent,nan2025nemori}. 
While these frameworks enable basic information storage and retrieval, agent memory demands more complex functionalities, such as relational modeling, hierarchical organization, and causal dependencies. Graph-based agent memory~\cite{zeng2024structural,jia2025aihippocampus} has emerged as the frontier for 2025–2026 research, transitioning from a passive `log' of facts to a structured topological model of experience that preserves how information is connected over time.
Unlike traditional linear or unstructured memory, graph-based memory can naturally encode relational dependencies between memory elements due to its intrinsic ability to model entity relationships, capture hierarchical semantics, and support flexible traversal and reasoning. Even plain memory can be regarded as a degenerate graph with trivial relationships, positioning graph-based agent memory as a general and flexible framework for agent memory design.
Recently, there has been a surge of research into graph-based memory architectures for LLM agents, including knowledge graphs (KG), temporal graphs, hypergraphs, hierarchical trees/graphs and hybrid graphs~\cite{sun2025hierarchical,rasmussen2025zep}, which demonstrated the efficacy across diverse scenarios, such as hierarchical task planning, multi-session conversational understanding, and neuro-symbolic reasoning.

Therefore, we present a comprehensive survey that consolidates the state-of-the-art in graph-based agent memory, categorizes their core techniques, synthesizes their applications, and identifies open challenges.
Our contributions are fourfold:
\begin{itemize}
    \item We propose a taxonomy of agent memory, including short-term vs. long-term memory, knowledge vs. experience memory, non-structural vs. structural memory, with an implementation view of graph-based memory (Section~\ref{sec:taxonomy}).
    \item We systematically analyze the critical memory management techniques, covering memory extraction (Section~\ref{sec:extraction}), memory storage (Section~\ref{sec:storing}), memory retrieval (Section~\ref{sec:retrieval}) and memory evolution (Section~\ref{sec:evolving}).
    \item We summarize open-sourced libraries and benchmarks (Section~\ref{sec:libraries_and_benchmarks}) that support the development and evaluation of self-evolving graph-based agent memory across diverse application scenarios (Section~\ref{sec:applications}).
    \item We identify critical challenges and outline future research directions to advance efficient and reliable graph-based agent memory systems (Section~\ref{sec:limitations}).
\end{itemize}

This survey aims to provide a holistic overview of graph-based agent memory, offering valuable insights for researchers to advance memory design and enabling practitioners to select appropriate structures and techniques for specific applications.




\section{Preliminaries}\label{sec:def}
\begin{figure*}[!t]
	\centering
     \includegraphics[width=1.0\linewidth]{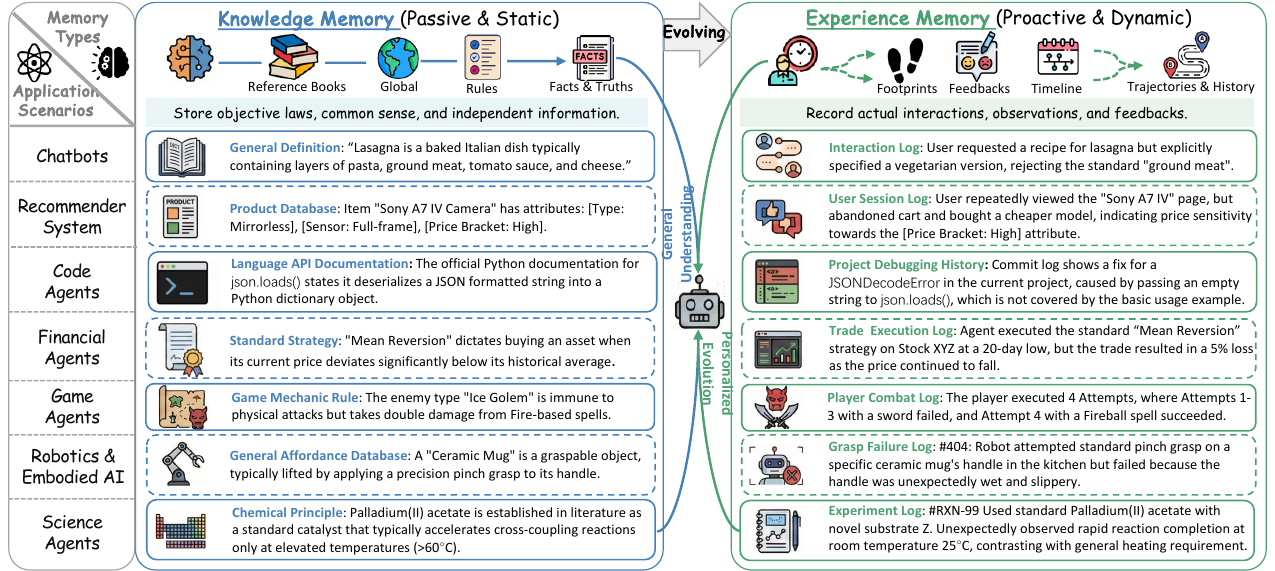}
        \caption{Two types of agent memory, i.e., \texttt{Knowledge Memory} and \texttt{Experience Memory}, and their application across different agent scenarios. The synergy between static knowledge memory and dynamic experience memory enables agents to both understand the world's rules and adapt to personal interactions.} 
        \label{fig:mem_type}
\end{figure*}

\begin{definition}[AI Agents]
\label{def:ai_agents}
An AI agent is an LLM-based system formed by four core modules:
\begin{itemize}
    \item \textbf{Perception Module}: sensing the environment and converting external observations into internal representations.
    \item \textbf{Reasoning Module}: decomposing complex tasks, reasoning about dependencies, interacting with memory, and formulating execution strategies.
    \item \textbf{Memory System}: comprising \textit{short-term memory} for immediate reasoning and \textit{long-term memory} for experience retention to support the reasoning.
    \item \textbf{Action Module}: executing actions in the environments.
\end{itemize}
The objective of the AI agent is to maximize the desired evaluation metrics, e.g., accuracies, success rates, and rewards.
\end{definition}

In this section, we present the preliminaries of AI agents. An AI agent is an LLM-based system to autonomously complete complex tasks by leveraging the LLM's capabilities for reasoning, memorizing and decision making. The formal definition is presented in Definition~\ref{def:ai_agents}. A typical execution paradigm of AI agents is perception-reasoning-action cycle, where the AI agent receives the inputs from the environment through its perception module, then reasoning with LLMs across the inputs, the LLM's internal knowledge and the contents stored in the memory system, and finally outputs the actions to execute into the environment. Within this paradigm, the memory systems play an important role for the AI agent's reasoning and action process. The basic operations of memory are defined in Definition~\ref{def:basic_operations}.

\begin{definition}[Basic Operations of Memory]
\label{def:basic_operations}
The basic operations of agent memory define the primitive actions that manipulate the memory $\mathcal{M}$. These atomic operations include:
\begin{itemize}
    \item \textbf{Write}: $\text{Write}(m, \mathcal{M}) \rightarrow \mathcal{M}'$, adding a new memory unit $m$ into the memory repository $\mathcal{M}$.
    \item \textbf{Read}: $\text{Read}(q, \mathcal{M}) \rightarrow \mathcal{M}_q$, retrieving relevant memory units $\mathcal{M}_q \subseteq \mathcal{M}$ based on query $q$.
    \item \textbf{Update}: $\text{Update}(m, \mathcal{M}) \rightarrow \mathcal{M}'$, modifying existing memory units based on new information.
    \item \textbf{Delete}: $\text{Delete}(m, \mathcal{M}) \rightarrow \mathcal{M}'$, removing obsolete or irrelevant memory units from $\mathcal{M}$.
\end{itemize}
\end{definition}
These operations collectively enable the dynamic management and evolution of the agent's memory system.

While the basic operations describe the atomic actions on memory, understanding how these operations are orchestrated over time is crucial for the full functionality of agent memory systems. Therefore, we further formalize the temporal dynamics of memory through the concept of lifecycle, which characterizes how information flows through different processing stages.

\begin{figure*}[!t]
	\centering
    \includegraphics[width=1.0\linewidth]{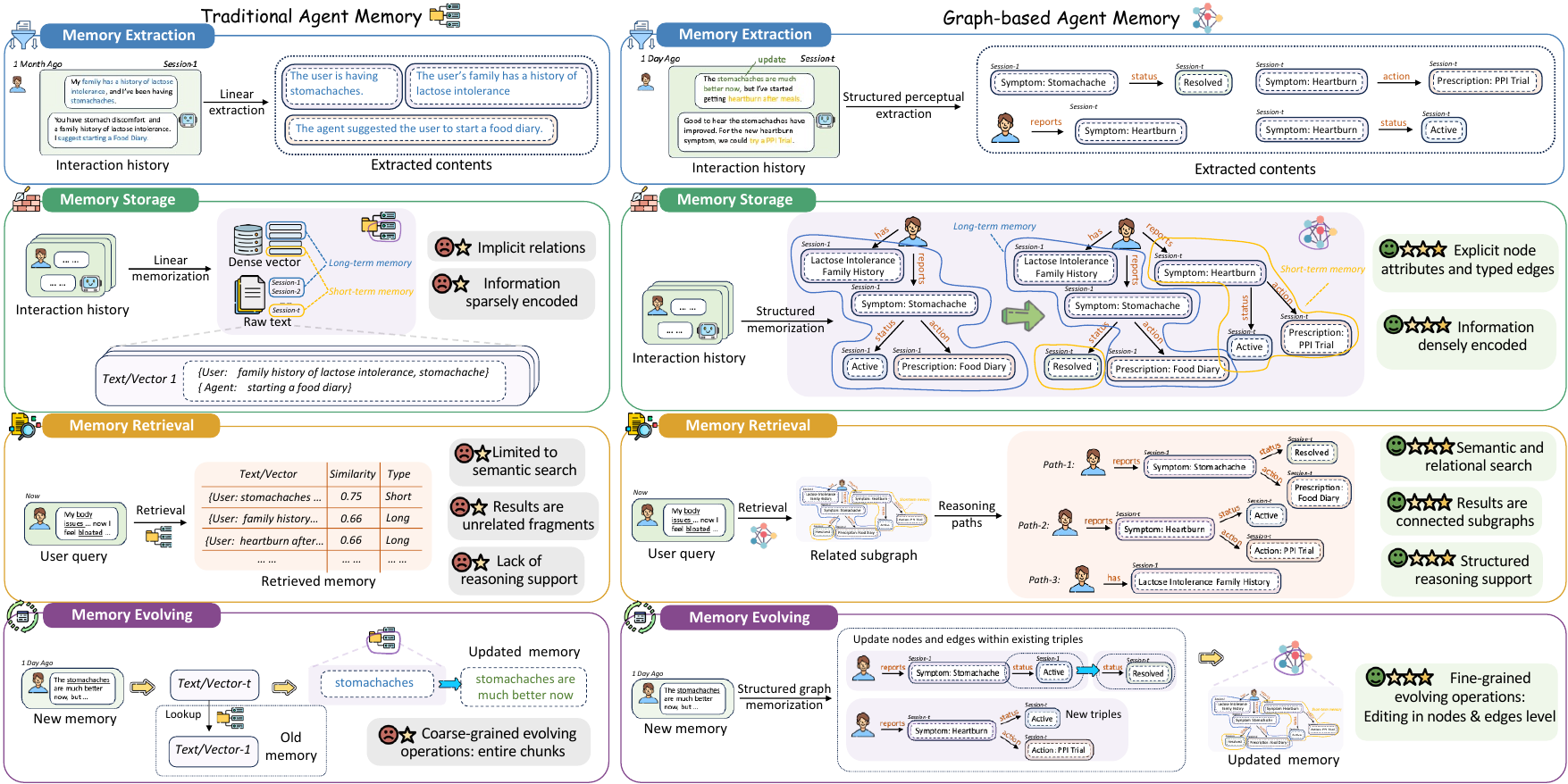}
    \caption{Comparison between traditional agent memory and graph-based agent memory} 
    \label{fig:comparison}
    \vspace{-10pt}
\end{figure*}


\begin{definition}[Lifecycle of Agent Memory]
The Lifecycle of Agent Memory describes the continuous, temporal flow of information processing within the agent, defining how raw data is transformed into stored knowledge and how the knowledge evolves over time. The lifecycle $\mathcal{L}$ is defined as a cyclic process consisting of four distinct stages:
\begin{itemize}
    \item \textbf{Memory Extraction}: The transformation of raw unstructured observations $o_t$  into structured memory units $m$.
    \item \textbf{Memory Storage}: The organization and placement of extracted units into the memory structure $\mathcal{M}$ through appropriate indexing and structural organization.
    \item \textbf{Memory Retrieval}: The mechanism of retrieving relevant stored information $\mathcal{M}_{\text{rel}} \subset \mathcal{M}$ in response to the query $q$.
    \item \textbf{Memory Evolution}: The post-processing phase where memory is refined, including internal self-evolving mechanisms (e.g., consolidation, abstraction) and external self-exploration processes (e.g., new environmental feedback).
\end{itemize}

\end{definition}
This lifecycle ensures that memory remains current, relevant, and optimally structured for supporting the agent's reasoning.

\section{Taxonomy of Agent Memory}
\label{sec:taxonomy}
In this section, we present a comprehensive taxonomy for agent memory from different perspectives. Specifically, memory is categorized on the basis of multiple dimensions, including temporal scope, functional roles and representational structures. We then introduce the graph-based memory as a unified view of agent memory and conclude this section from an implementation perspective.


\subsection{Long-term vs. Short-term Memory}
One of the most straightforward categorization of memory is the temporal dimension, i.e., information retention durations:
\begin{itemize}
    \item \textbf{Short-term Memory}: Maintains recent, immediately relevant information with rapid access and limited capacity. This includes the current conversation context, active reasoning traces, and transient state variables. Short-term memory is volatile and typically discarded after the immediate task completion, though significant elements may be consolidated into long-term storage.
    \item \textbf{Long-term Memory}: Stores persistent information across sessions, including accumulated knowledge, historical interactions, learned patterns, and user preferences. Long-term memory supports continuity across episodes, enables transfer learning, and provides the foundation for personalization and adaptive behavior over extended timescales.
\end{itemize}

\subsection{Cognitive Structure of Memory}

We introduce the cognitive structure of the memory in~\cite{sumers2023cognitive}:
\begin{itemize}[leftmargin=*, topsep=0pt, partopsep=0pt, parsep=0pt, itemsep=2pt]
    \item \textbf{Semantic Memory}: Stores general, decontextualized world knowledge and factual information (e.g., ``Paris is the capital of France''). Within the graph structure, this forms the stable ontology, providing common sense and domain-specific facts that ground the agent's reasoning.
    
    \item \textbf{Procedural Memory}: Encodes skills, routines, and immutable rules. It represents ``how-to'' knowledge, such as standard operating procedures or game rules. In application, this allows agents to execute complex tasks automatically under standard conditions.
    
    \item \textbf{Associative Memory}: Establishes latent links between different concepts or data points within the knowledge base. By connecting related pieces of information (e.g., symptoms to diagnoses, or products to categories), it facilitates creative reasoning and analogy-making, acting as the connective tissue of the knowledge graph.

    \item \textbf{Working Memory}: Acts as the agent's ``mental scratchpad'' for immediate experience. It temporarily holds the current conversation turn, intermediate reasoning steps, and transient variables. While ephemeral, it is the entry point for all experience memory, characterized by fast access and direct influence on the immediate next action.
    
    \item \textbf{Episodic Memory}: Records the chronological sequence of past sessions. It transforms transient working memory into a persistent and queryable autobiographical history. For instance, it notes that a customer requested a vegetarian option or a canceled order. This allows the agent to recall ``what happened and when,'' providing temporal grounding.
    
    \item \textbf{Sentiment Memory}: Captures the emotional tone or sentiment derived from interactions. By logging user feedback or frustration levels, the agent can adapt its empathy and style in future turns. This layer adds a qualitative dimension to the raw interaction logs.
\end{itemize}

\subsection{Knowledge vs. Experience Memory}

To define agent memory concretely, we can draw a parallel to the human cognitive system. Human memory is a structured process involving the encoding, storage, and retrieval of information, often categorized into different types. Similarly, agent memory can be understood through two primary, complementary categories as shown in Figure~\ref{fig:mem_type}:

\noindent\textbf{Knowledge Memory (Passive \& Static)}: This represents the agent's passive and static repository of objective, global, and verifiable information. It functions as an internal reference library or textbook, containing canonical facts, universal rules, established procedures, and general truths about the world. This memory is typically pre-loaded, slowly updated, and context-independent. Its purpose is to provide a stable, reliable foundation for reasoning and action. For instance, it stores the factual definition of concepts, the immutable rules of a game, or standard scientific principles. In applications, a shopping agent's product database or a robot's general affordance model constitutes Knowledge Memory. Generally, this memory is passive, serving as a reliable, factual backbone for reasoning.

\noindent\textbf{Experience Memory (Proactive \& Dynamic)}: This is the agent's personal logbook, actively recording its specific interactions, observations, and the outcomes of its actions. It includes user dialogue history, execution logs, trial-and-error trajectories, and feedback. For instance, it notes that a particular user requested a vegetarian lasagna variant, that a trade based on a mean-reversion strategy actually resulted in a loss, or that a standard grasping procedure failed on a wet mug. This memory is dynamic, personalized, and forms the basis for learning from practice and adapting to specific contexts.

\subsection{Non-structural vs. Structural Memory} 
Traditional agent memory systems commonly adopt simple storage paradigms, including:

\begin{itemize}[leftmargin=*, topsep=0pt, partopsep=0pt, parsep=0pt, itemsep=0pt]
    \item \textbf{Linear or buffer-based memory}, such as fixed-length token windows or conversation histories, which maintain recent interactions but suffer from information loss and lack relational context.
    \item \textbf{Vector-based memory}, which encodes experiences into dense embeddings stored in vector databases, enabling semantic similarity search but struggling with structured reasoning and hierarchical relationships.
    \item \textbf{Key-value or log-based memory}, which records events in sequential logs or attribute-value pairs, supporting straightforward lookup but offering limited capability for complex querying or dynamic updates.
\end{itemize}

A common thread among these traditional paradigms is their treatment of memory as a sequential, flat, or implicitly structured store. While effective for certain patterns, they often fall short in explicitly representing and efficiently reasoning over the complex web of relationships between pieces of knowledge, a capability critical for sophisticated planning, causal understanding, and narrative coherence. More importantly, while these approaches enable basic recall and short-term context management, they exhibit notable limitations in scenarios requiring hierarchical knowledge organization, temporal tracking, and long-term adaptive learning. These constraints become particularly evident in long-horizon tasks, multi-session interactions, and domains where knowledge evolves dynamically.

\subsection{Graph-based Memory: A Unified  and General Perspective}

Graph-based agent memory emerges as a powerful generalization and enhancement of conventional memory frameworks. The core idea of graph-based agent memory is modeling memory content as a dynamic, structured \textbf{Memory Graph}. In this paradigm, memory units (e.g., events, entities, concepts, observations) are abstracted as \textbf{nodes}, and the semantic, temporal, causal, or logical relationships between them are abstracted as \textbf{edges}. This explicit structural representation transforms memory from a flat list of entries or a hidden state vector into a rich, interconnected network of knowledge.

By representing memory elements as nodes and their relationships as edges, graph structures naturally support: \ding{182} Explicit relationship modeling, allowing agents to reason over causal dependencies and semantic associations between memory items; \ding{183} Hierarchical organization, from fine-grained factual triples to general thematic clusters or subgraphs; \ding{184} Temporal and dynamic structuring, where time-aware edges can capture event sequences, state transitions, and knowledge evolution; \ding{185} Efficient structured retrieval, enabling traversal, subgraph extraction, and multi-hop relational queries beyond mere semantic similarity.

Notably, traditional memory forms can be viewed as degenerate or simplified cases within the graph memory paradigm. For instance, a linear buffer corresponds to a chain within a graph, and a vector memory can be interpreted as a fully-connected graph with similarity-weighted edges. Thus, graph-based memory does not merely replace existing designs but provides a unified and extensible framework.

In essence, graph-based agent memory elevates memory from a passive, flat ``log" to an active, structured ``knowledge graph" tailored to the agent's lived experience. It not only records ``what happened" but, more importantly, models ``how these things are connected." This provides a powerful foundation for associative reasoning, modeling long-term dependencies, and achieving explainable agent behavior. Figure~\ref{fig:comparison} summarizes the key distinctions between traditional agent memory and graph-based agent memory. In summary, graph-based agent memory re-conceptualizes memory from a passive recording into an active, structured model of experience. By making relationships first-class citizens, it provides a powerful foundation for the complex reasoning, long-term coherence, and adaptive behavior required by sophisticated autonomous agents. Graph-based memory architectures, such as knowledge graphs and hypergraphs, have demonstrated superior performance in applications requiring multi-session coherence, personalized adaptation, complex task planning, and hallucination reduction. The subsequent sections delve into the technical realization of such memory systems, covering construction, retrieval, updating, and domain-specific applications.

\subsection{An Alternative View from Implementation}

In this paper, we use the lifecycle to introduce the memory, however, we can also have an alternative view of the memory from the implementation perspective, which can help readers to understand this paper as a researcher or engineer. Specifically, the framework begins with underlying structures, which encompasses the foundational data representations including graph-based structures, embeddings, and temporal sequences that form the basis of memory storage (Section~\ref{sec:storing}). The stored representations manifest as various memory contents, including conversational history, temporal records, and structured knowledge bases. These contents are curated by memory processes of extraction (Section~\ref{sec:extraction}) and evolution (Section~\ref{sec:evolving}), where relevant information is filtered and refined based on context and usage patterns. Finally, the usage interface enables Retrieval (Section~\ref{sec:retrieval}) of relevant information and facilitates reasoning tasks by accessing this curated memory. This implementation view provides a systematic perspective from the implementation on how language models maintain, process, and leverage memory to support the complex reasoning.

\definecolor{colorStoring}{HTML}{E8A6B1} 
\definecolor{colorRetrieval}{HTML}{C9BF8E} 
\definecolor{colorStrategy}{HTML}{CCD376} 
\definecolor{colorExtraction}{HTML}{4A9DAF} 
\definecolor{colorEvolving}{HTML}{A28CC2} 

\tikzstyle{my-box}=[
    rectangle,
    draw=hidden-draw,
    rounded corners,
    align=left,
    text opacity=1,
    minimum height=0.5em,
    minimum width=4em,
    fill opacity=.8,
]

\tikzstyle{leaf-head}=[my-box,
    draw=gray,
    fill=gray!10, 
    text=black,
    font=\scriptsize,
    inner xsep=1pt,
    inner ysep=1pt,
    line width=0pt,
    align=left,
]

\tikzstyle{node-Extraction}=[my-box, draw=colorExtraction, fill=colorExtraction!25, text=black, font=\scriptsize, inner xsep=1pt, inner ysep=1pt, line width=0pt]
\tikzstyle{node-Storing}=[my-box, draw=colorStoring, fill=colorStoring!25, text=black, font=\scriptsize, inner xsep=1pt, inner ysep=1pt, line width=0pt]
\tikzstyle{node-Retrieval}=[my-box, draw=colorRetrieval, fill=colorRetrieval!25, text=black, font=\scriptsize, inner xsep=1pt, inner ysep=1pt, line width=0pt]
\tikzstyle{node-Operator}=[my-box, draw=colorOperator, fill=colorOperator!25, text=black, font=\scriptsize, inner xsep=1pt, inner ysep=1pt, line width=0pt]
\tikzstyle{node-Strategy}=[my-box, draw=colorStrategy, fill=colorStrategy!25, text=black, font=\scriptsize, inner xsep=1pt, inner ysep=1pt, line width=0pt]
\tikzstyle{node-Evolving}=[my-box, draw=colorEvolving, fill=colorEvolving!25, text=black, font=\scriptsize, inner xsep=1pt, inner ysep=1pt, line width=0pt]

\tikzstyle{lnode-Extraction}=[my-box, minimum height=1.2em, draw=colorExtraction, fill=white, text=black, font=\scriptsize, inner xsep=1pt, inner ysep=1pt, line width=0pt, text width=26em]
\tikzstyle{lnode-Storing}=[my-box, minimum height=1.2em, draw=colorStoring, fill=white, text=black, font=\scriptsize, inner xsep=1pt, inner ysep=1pt, line width=0pt, text width=26em]
\tikzstyle{lnode-Retrieval}=[my-box, minimum height=1.2em, draw=colorRetrieval, fill=white, text=black, font=\scriptsize, inner xsep=1pt, inner ysep=1pt, line width=0pt, text width=26em]
\tikzstyle{lnode-Evolving}=[my-box, minimum height=1.2em, draw=colorEvolving, fill=white, text=black, font=\scriptsize, inner xsep=1pt, inner ysep=1pt, line width=0pt, text width=26em]

\begin{figure*}
    \centering
    \resizebox{1.\textwidth}{!}{
        \begin{forest}
            forked edges,
            for tree={
                grow=east,
                reversed=true,
                anchor=base west,
                parent anchor=east,
                child anchor=west,
                base=left,
                font=\scriptsize,
                rectangle,
                draw=hidden-draw,
                rounded corners,
                align=left,
                minimum width=1em,
                edge+={darkgray!40, line width=0.6pt},
                s sep=3pt,
                inner xsep=1pt,
                inner ysep=2pt,
                line width=0pt,
                ver/.style={rotate=90, child anchor=north, parent anchor=south, anchor=center},
            }, 
            [{Graph-based Agent Memory}, leaf-head, ver
                [{Extraction (\S\ref{sec:extraction})},node-Extraction, text width=6em
                    [Textual Data, node-Extraction, text width=7.5em
                        [{LLM-Induce-Graph\cite{yousuf2025llminducegraphinvestigating}, Structural Memory\cite{Zeng2024OnTS}, PersonaAgent\cite{Liang2025PersonaAgentWG},  G-CoT\cite{Huan2025ScalingGC},\\HiAgent\cite{Hu2024HiAgentHW}}, lnode-Extraction]
                    ]
                    [Sequential Data, node-Extraction, text width=7.5em
                        [{Reflexion~\cite{Shinn2023ReflexionAA}, Mem-$\alpha$~\cite{wang2025mem}}, lnode-Extraction]
                    ]
                    [Multimodal Data, node-Extraction, text width=7.5em
                        [{MemoryVLA~\cite{Shi2025MemoryVLAPM}, Multi-Temporal~\cite{Yeo2023MultiTemporalLM}, Optimus-1~\cite{li2024optimus}}, lnode-Extraction]
                    ]
                ]
                [{Storage (\S\ref{sec:storing})}, node-Storing, text width=6em
                    [Knowledge Graph, node-Storing, text width=7.5em
                        [{MemLLM~\cite{modarressi2024memllm}, AriGraph~\cite{anokhin2024arigraph}, Mem0~\cite{chhikara2025mem0}}, lnode-Storing]
                    ]
                    [Hierarchical Structure, node-Storing, text width=7.5em
                        [{DAMCS~\cite{yang2025llm}, G-Memory~\cite{zhang2025g}, ENGRAM~\cite{patel2025engram}, SGMem~\cite{wu2025sgmem}, Personalized Agents~\cite{wang2024crafting}}, lnode-Storing]
                    ]
                    [Temporal Graph, node-Storing, text width=7.5em
                        [{Zep~\cite{rasmussen2025zep}, TReMu~\cite{ge2025tremu}, MemoTime~\cite{tan2025memotime}}, lnode-Storing]
                    ]
                    [Hypergraph Structure, node-Storing, text width=7.5em
                        [{HyperGraphRAG~\cite{luo2025hypergraphrag}, HyperG~\cite{huang2025hyperg}}, lnode-Storing]
                    ]
                    [Hybrid Architectures, node-Storing, text width=7.5em
                        [{Optimus-1~\cite{li2024optimus}, KG-Agent~\cite{jiang2025kg}}, lnode-Storing]
                    ]
                ]
                [{Retrieval (\S\ref{sec:retrieval})}, node-Retrieval, text width=6em
                    [Similarity-based, node-Retrieval, text width=7.5em
                        [{Multimodal Agent~\cite{long2025seeing},  Zep~\cite{rasmussen2025zep}, Mem0~\cite{chhikara2025mem0}, G-Memory~\cite{zhang2025g}}, lnode-Retrieval]
                    ]
                    [Rule-based, node-Retrieval, text width=7.5em
                        [{MemInsight~\cite{salama2025meminsight}, FinMem~\cite{yu2025finmem},  MyAgent~\cite{hou2024my}, Neural Graph Memory~\cite{fisher2025neural}, TReMu~\cite{ge2025tremu}}, lnode-Retrieval]
                    ]
                    [Graph-based, node-Retrieval, text width=7.5em
                        [{Mem0~\cite{chhikara2025mem0}, Zep~\cite{rasmussen2025zep}, SimGRAG~\cite{cai2025simgrag}, CAN~\cite{wu2024can}, SGMem~\cite{wu2025sgmem}, H-MEM~\cite{h-mem}, \\ G-Memory~\cite{zhang2025g}, KG-Agent~\cite{jiang2025kg}}, lnode-Retrieval]
                    ]
                    [Temporal-based, node-Retrieval, text width=7.5em
                        [{MemoTime~\cite{tan2025memotime}, Mnemosyne~\cite{jonelagadda2025mnemosyne}, AssoMem~\cite{zhang2025assomem}, Zep~\cite{rasmussen2025zep}, LiCoMemory~\cite{huang2025licomemory}}, lnode-Retrieval]
                    ]
                    [RL-based, node-Retrieval, text width=7.5em
                        [{Personalized Agents~\cite{wang2024crafting}, Memento~\cite{zhou2025memento}, Mem-$\alpha$~\cite{wang2025mem}, Memory-R1~\cite{yan2025memory}, TGM\cite{xia2025experience}, \\ MAQR~\cite{xu2025memory}, Reflective Memory~\cite{tan2025prospect}}, lnode-Retrieval]
                    ]
                    [Agent-based, node-Retrieval, text width=7.5em
                        [{KG-HLM~\cite{kim2024leveraging},  GraphCogent~\cite{wang2025graphcogent},  Omni Memory~\cite{wang2025omni}, Optimus-1~\cite{li2024optimus},\\  Cradle~\cite{tan2025cradle}, DAMCS~\cite{yang2025llm}, MemGPT~\cite{packer2023memgpt}, CoALA~\cite{sumers2023cognitive}, HiAgent~\cite{hu2025hiagent}, AriGraph~\cite{anokhin2024arigraph},\\ MAS-RL~\cite{jia2025enhancing}, Collabor Memory~\cite{rezazadeh2025collaborative}}, lnode-Retrieval]
                    ]
                    [Multi-round, node-Retrieval, text width=7.5em
                        [{Deep Research~\cite{yan2025general}, Assistant Memory~\cite{zhang2025bridging}, RCR-Router~\cite{liu2025rcr}, MemSearcher~\cite{yuan2025memsearcher}, \\ MemoTime~\cite{tan2025memotime},  GITM~\cite{zhu2023ghost},  LEGOMem~\cite{han2025legomem} }, lnode-Retrieval]
                    ]
                    [Post-retrieval, node-Retrieval, text width=7.5em
                        [{SimGRAG~\cite{cai2025simgrag}, Mirix~\cite{wang2025mirix}, MemGen~\cite{zhang2025memgen}}, lnode-Retrieval]
                    ]
                    [Hybrid-retreival, node-Retrieval, text width=7.5em
                        [{MEM1~\cite{zhou2025mem1}, MemSearcher~\cite{yuan2025memsearcher}}, lnode-Retrieval]
                    ]
                ]
                [{Evolution (\S\ref{sec:evolving})}, node-Evolving, text width=6em
                    [Internal Self-Evolving, node-Evolving, text width=7.5em
                        [{Zep~\cite{rasmussen2025zep}, Nemori~\cite{nan2025nemori}, Mem-$\alpha$~\cite{wang2025mem}, GraphRAG~\cite{edge2024local}, RecallM~\cite{kynoch2023recallmadaptablememorymechanism}, Agent KB~\cite{tang2025agent},\\ Mem0~\cite{chhikara2025mem0}, FLEX~\cite{cai2025flex}, Reflexion~\cite{Shinn2023ReflexionAA}, ToG~\cite{sun2024thinkongraph}, RRP~\cite{RRP}, MemoryBank~\cite{zhong2024memorybank},\\ RoG~\cite{luo2024reasoning},  Memory OS~\cite{kang2025memory}, MemGPT~\cite{packer2023memgpt}}, lnode-Evolving]
                    ]
                    [External Self-Exploration, node-Evolving, text width=7.5em
                        [{MATRIX~\cite{xu2024matrix}, Memory-R1~\cite{yan2025memory}, Inside-Out~\cite{gekhman2025inside}, MemEvolve~\cite{zhang2025memevolve}, Kimi K2.5~\cite{moonshot2026k25},\\ ExpeL~\cite{zhao2024expel},  Proactive Memory~\cite{yang2026beyond}, AgentEvolver~\cite{zhai2025agentevolver}}, lnode-Evolving]
                    ]
                ]
            ]
        \end{forest}}
    \caption{A Comprehensive Taxonomy of Graph-based Memory Management for LLM Agents.}
    \label{fig:taxonomy}
\end{figure*}
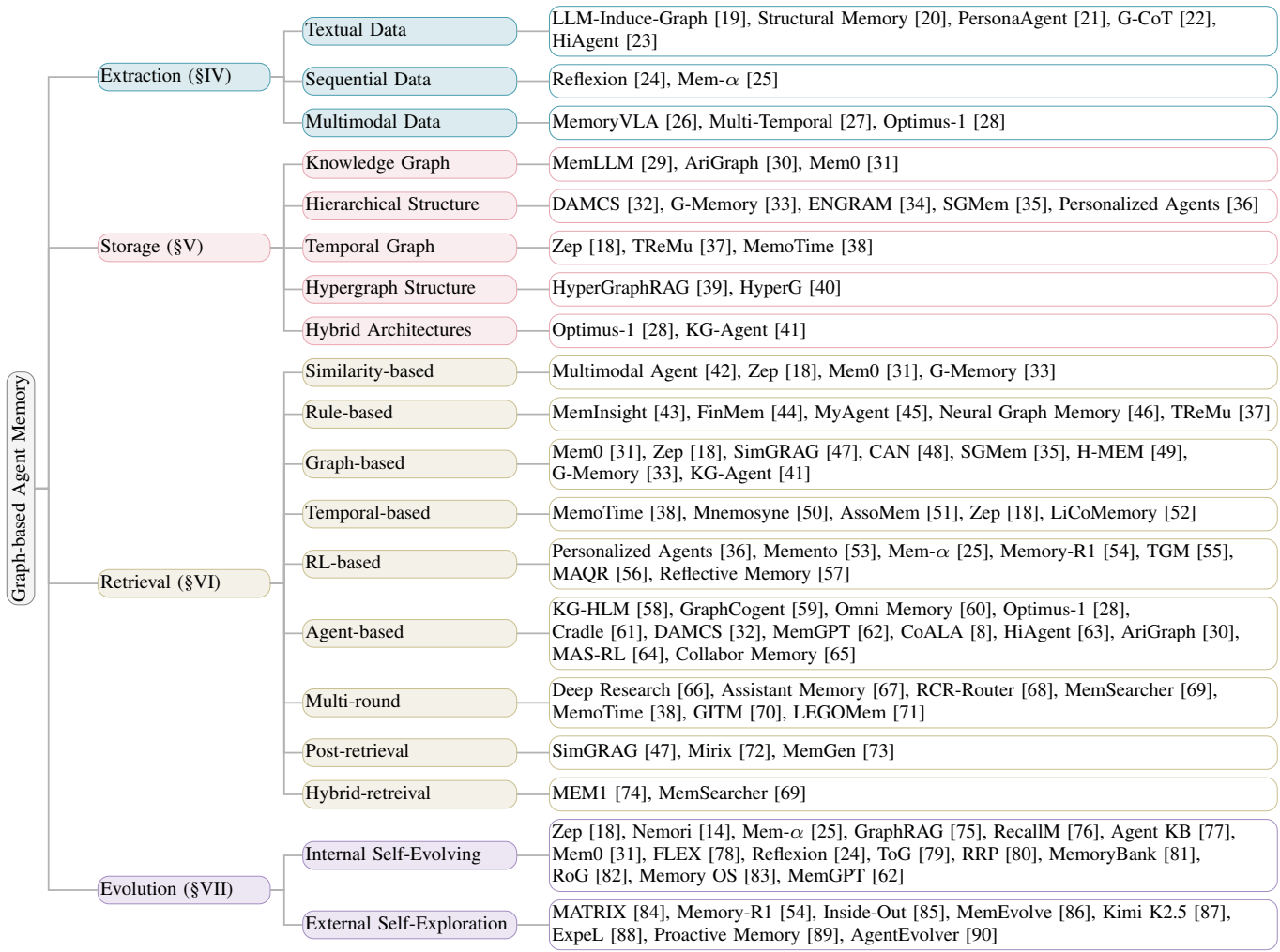

\section{Memory Extraction: Transforming the Data}
\label{sec:extraction}
\begin{figure*}[h]
	\centering
        \includegraphics[width=1.0\linewidth]{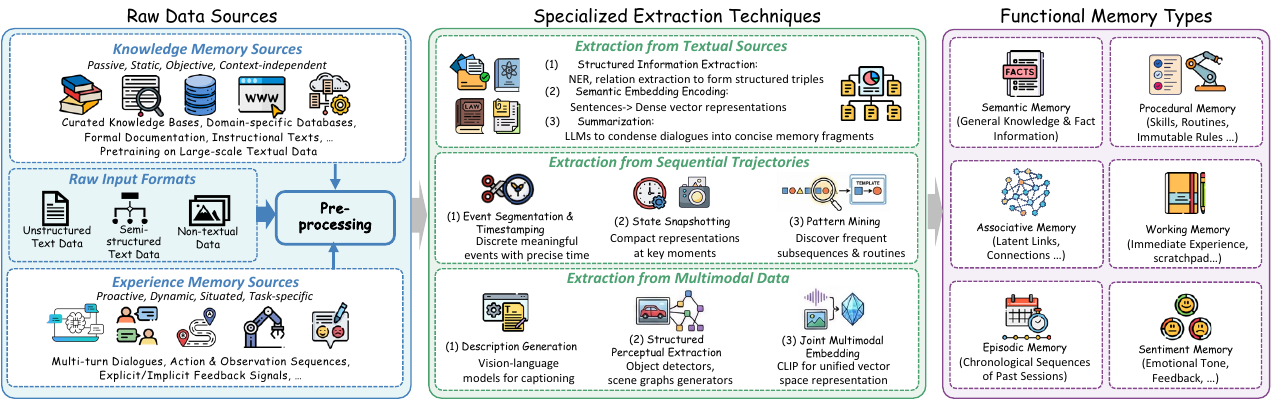}
        \caption{Overview of agent memory extraction. This figure illustrates a unified pipeline of agent memory construction from various data resources. Raw inputs, originating from both experience and knowledge memory, are transformed into structured and compact representations through specialized extraction techniques. These extracted units are then organized into distinct functional memory types, enabling agents to support reasoning and downstream tasks.} 
        \label{fig: illustration_memory_extraction}
\end{figure*}


The process of memory extraction begins with the collection and pre-processing of raw data. These raw inputs, which vary widely in format and constitute the foundational material from which an agent’s internal memories are constructed. From this perspective, memory extraction can be understood along two complementary dimensions: the source from which memory is derived and the form in which the extracted memory is ultimately encoded. Based on the definitions introduced in the previous section, agent memory can be broadly categorized into knowledge and experience memory, which differ primarily in their sources and roles within the system.

\textbf{Experience memory} is extracted from the agent’s own interaction history and reflects the accumulation of situated, task-specific experience over time. Its primary sources include multi-turn dialogues with users, action and observation sequences produced during task execution, and explicit or implicit feedback signals. 
\textbf{Knowledge memory}, in contrast, is extracted from sources that exist independently of the agent’s personal experience and are intended to represent objective and generally valid information. Typical sources include curated knowledge bases, domain-specific databases, formal documentation, instructional texts, and other authoritative corpora. In addition, a substantial portion of knowledge memory is implicitly acquired through pretraining on large-scale textual data, where factual and procedural information becomes embedded in the model parameters. Information drawn from these sources is usually stable, context-independent, and broadly applicable. Consequently, knowledge memory extraction focuses on distilling canonical facts, rules, and procedures into persistent representations that provide a reliable foundation for reasoning across tasks and domains.

The underlying information shows up in many different formats. These sources may appear as unstructured or semi-structured text such as documents, dialogue transcripts, and execution logs. Others may be non-textual, including images, sensory observations, or structured records generated during interaction. This diversity in source modality means that raw information cannot be directly stored as memory. Instead, it must be processed and abstracted in ways that reflect both the characteristics of the source and the type of memory being constructed. As a result, different sources naturally require different extraction strategies. The following outlines the primary techniques used for different types of data sources.

\paragraph{Extraction from Textual Sources} The extraction of memory from unstructured textual data, such as conversational logs or documents, focuses on identifying and structuring semantic information~\cite{yousuf2025llminducegraphinvestigating}. Key approaches include: (1) Structured information extraction: Using named entity recognition and relation extraction models, or prompting large language models, to directly extract entities, attributes, and their relationships in the form of (entity-relation-entity) triples~\cite{Zeng2024OnTS,Liang2025PersonaAgentWG}. (2) Semantic embedding encoding: Encoding sentences or paragraphs into dense vector representations using models such as Sentence-BERT, transforming semantic content into embeddings suitable for similarity-based retrieval~\cite{Huan2025ScalingGC}. (3) Summarization: Applying extractive or abstractive summarization models, including large language models, to condense lengthy texts or dialogue histories into concise and informative memory fragments~\cite{Hu2024HiAgentHW}.

\paragraph{Extraction from Sequential Trajectories} For time-series interaction data, such as action-observation sequences, extraction aims to capture temporal structure: (1) Event segmentation and timestamping: Segmenting continuous trajectories into discrete, meaningful events or episodes and annotating each with precise timestamps~\cite{Shinn2023ReflexionAA}. (2) Dynamic State Snapshots: Beyond static snapshots, this method captures the evolution of critical states. It involves periodically capturing and storing compact representations of the agent’s or environment’s state at key moments, such as embeddings or feature vectors~\cite{wang2025mem}.  (3) Pattern mining: Applying offline sequence mining or clustering algorithms to discover frequently occurring successive sub-sequences or strategic routines, which are then abstracted into procedural memory templates.

\paragraph{Extraction from Multimodal Data}
For sensory data such as vision or audio, extraction bridges the gap between raw signals and semantic meaning:
(1) Description generation: Using vision-language models~\cite{Shi2025MemoryVLAPM} or audio captioning models~\cite{Yeo2023MultiTemporalLM} to generate textual descriptions of visual scenes or auditory events, which are subsequently processed as text. (2) Interactive content: This process focuses on detecting objects or distilling information between an agent’s actions and environmental responses such as raw pixels or signals. (3) Joint multimodal embedding: Encoding data from different modalities into a unified vector space using customized models, producing a joint embedding that serves as a compact memory representation of the multimodal experience~\cite{li2024optimus}.

As shown in Figure \ref{fig: illustration_memory_extraction}, the raw data is transformed into structured and semantically enriched representations by the extraction process. This transformation typically occurs across three levels of abstraction: from the original raw data flows, to intermediary extracted entities and relationships, and finally into organized functional memory types that serve specific cognitive and operational purposes within the agent, which is introduced in Section~\ref{sec:taxonomy}.

\section{Memory Storage: Organizing the Mind} 
\label{sec:storing}

The extraction stage described above produces a set of semantically enriched artifacts, including identified entities and relations, dense semantic embeddings, concise summaries, segmented events with timestamps, and multimodal captions, which together form the operative substrate for downstream memory architectures. A central challenge in building agent memories is therefore to transform these heterogeneous extracted artifacts into storage formats that preserve relevant semantics while supporting efficient retrieval and reliable updates. Unlike static knowledge bases, agent memories must also accommodate dynamics, personalization and experience grounding, all of which shape how extracted information should be encoded and maintained.

Choosing a particular graph based memory structure amounts to explicit trade offs between competing design goals. Precision and explicit multiple hop reasoning favor relational graphs; compression and conceptual abstraction favor hierarchical or treelike summaries; temporal fidelity motivates temporal knowledge graphs and time indexed episodes; and cross modal generalization or fuzzy recall often favors vector stores or hybrid systems. With these trade offs in mind, this section focuses first on Knowledge Graphs as a canonical relational substrate, explaining how triples are produced, integrated and maintained, and then surveys hierarchical, temporal, hypergraph and hybrid architectures that address complementary points along this design spectrum. Figure~\ref{fig:taxnomony} summarizes the construction paradigms discussed below.

\subsection{Knowledge Graph Structure}
\label{subsec:kg_as_memory}

The Knowledge Graph (KG) stands as a structured memory paradigm, explicitly designed to store and reason over factual knowledge~\cite{modarressi2024memllm}. It represents information as a network of interconnected \textit{triples}, where each triple takes the form (\textit{head entity}, \textit{relation}, \textit{tail entity}). This relational structuring makes it a powerful substrate for specific types of agent memory.

\noindent\paragraph{KG Modeling} Constructing a KG for an agent involves the continuous extraction and integration of structured triples from unstructured interaction streams. The primary method relies on Large Language Models serving as a powerful, open-vocabulary parsing engine. For instance, in the AriGraph world model \cite{anokhin2024arigraph}, the LLM parses each textual observation from the environment to identify relevant objects and extract their relationships in the form of triplets. Similarly, systems like Mem0~\cite{chhikara2025mem0} employ an LLM in their extraction phase to convert conversation messages into entity-relation triplets. This approach outperforms traditional Named Entity Recognition (NER) and relation classification pipelines, as LLMs can generalize to novel fine-grained entity types and exploit implicit relations based on contextual understanding.

The extracted triples then enter an update phase, where they are integrated into the persistent graph store. This phase is non-trivial, involving operations such as conflict detection (e.g., handling contradictory facts), relationship pruning, and schema evolution. Mem0 explicitly models this through a dedicated update mechanism that evaluates new memories against similar existing ones. This two-step process, LLM-based extraction followed by reasoned integration, forms the core pipeline for building a dynamic, experience-driven knowledge graph from an agent's raw perceptions.

\noindent\paragraph{Corresponding Memory Types} The triple-based structure of a KG makes it exceptionally suitable for implementing long-term static memory types. This includes general world facts (e.g., \textit{(Paris, Capital\_Of, France)}) and immutable domain-specific concepts. The explicit relational format allows for efficient storage and complex, multi-hop queries that are difficult for vector-based retrievers. Furthermore, by augmenting triples with temporal metadata or linking them to episodic events, where episodic vertices are connected to triples extracted from the same observation, KGs can also support episodic memory.



\subsection{Hierarchical Memory Structure}
\label{subsec:hierarchical_structure}

The hierarchical structure is the most common and intuitive paradigm for organizing knowledge within agent memory systems~\cite{yang2025llm,zhang2025g,wang2024crafting}. By arranging information into multi-level trees, it provides a rationale hierarchy to compress extensive experiences into manageable schema. The essence of a tree is a directed acyclic graph (DAG) that explicitly models \emph{parent-child} and \emph{containment} relationships, allowing it to represent concepts ranging from broad categories to specific instances and from high-level goals to granular execution steps. This inherent property enables efficient top-down navigation for querying abstract themes and bottom-up summarization for maintaining coherence across vast memory stores. Systems like MemTree adopt this approach by dynamically routing new information through the hierarchy, clustering similar content under existing nodes while creating new branches for novel information, with all ancestor nodes recursively updating their semantic summaries to reflect the integrated knowledge.

The construction of hierarchical memory mainly relies on two processes: \textbf{semantic clustering} for organization and \textbf{recursive summarization} for abstraction~\cite{patel2025engram}. To maintain the hierarchy's value as a compressed knowledge schema, each parent node dynamically synthesizes a concise linguistic summary of all information contained within its subtree. Beyond static factual knowledge mentioned above, such structure also manifests as \emph{Session or Sequence Memory}. Here, individual sentences or dialogue turns, are linked primarily by their temporal order or conversational flow within a defined session. This creates a timeline or narrative chain that is essential for modeling dialogue coherence and user intent evolution over a single interaction episode. The SGMEM ~\cite{wu2025sgmem} system proposes a sentence-level tree for long-term conversational agents, using the sequential and referential links between utterances to construct a coherent interaction.


\begin{figure*}[t]
    \centering
    \includegraphics[width=\linewidth]{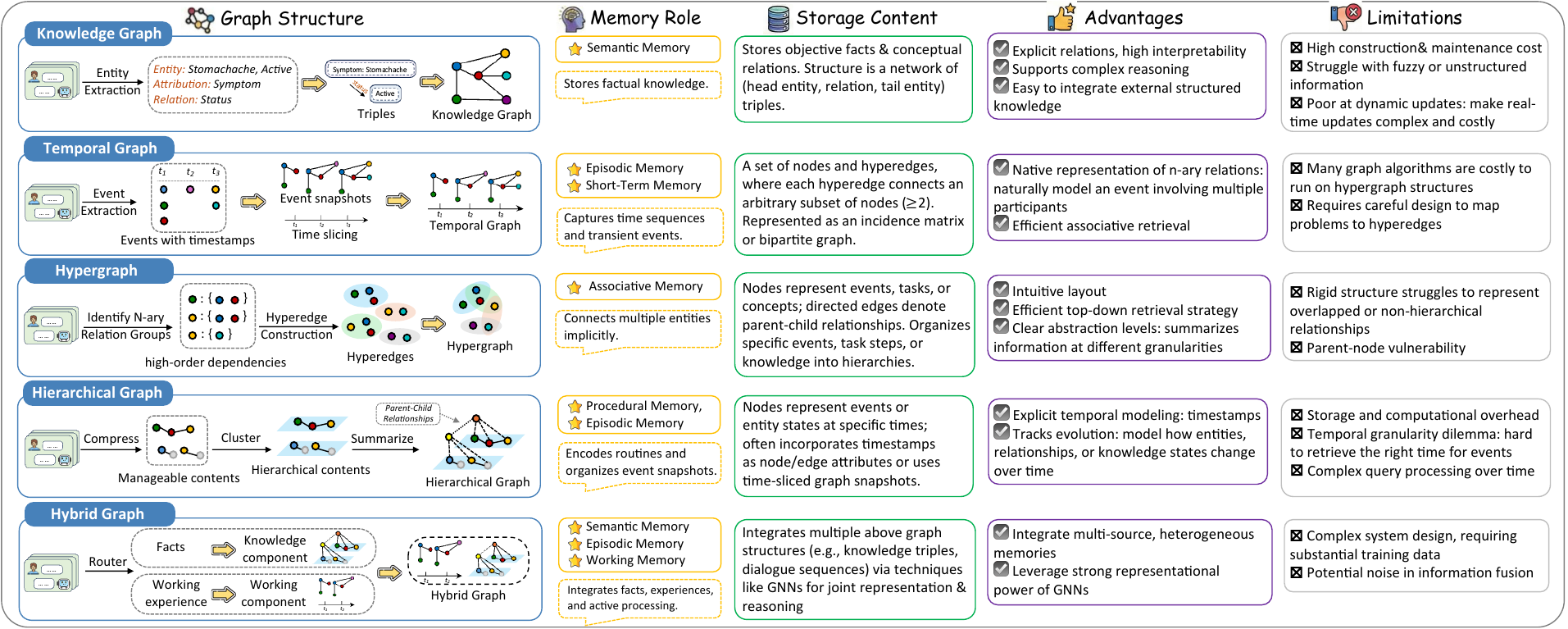}
    \caption{A Comprehensive Taxonomy of Graph Construction Paradigms for Agent Memory Systems: Methodologies, Corresponding Memory Functions, and Comparative Analysis of Advantages and Limitations.}
    \label{fig:taxnomony}
\end{figure*}

\subsection{Temporal Graph Structure}
\label{sec:temporal_graph}
Real-world interactions are inherently dynamic, where facts hold validity only within specific time windows. Temporal Knowledge Graphs (TKGs) extend standard triples to quadruples $(s, r, o, t)$, but simple timestamping is often insufficient for complex agentic workflows. Recent architectures have introduced more granular temporal modeling to handle validity, ambiguity, and reasoning monotonicity.

\paragraph{Bi-Temporal Modeling}
A critical distinction in agent memory is between the time an event occurs (\textit{valid time}) and the time it is recorded (\textit{transaction time}). \textbf{Graphiti}~\cite{rasmussen2025zep} implements a bi-temporal model tracking two distinct timelines. This allows the system to manage evolving conversations where a fact introduced at $t_1$ might retrospectively describe an event at $t_2$. By explicitly tracking creation ($t'_{created}$) and expiration ($t'_{expired}$) timestamps alongside validity intervals, the system can resolve contradictions through temporal invalidation rather than overwrites, maintaining a faithful history of state changes.

\paragraph{Disentangling Mention and Event Time}
In multi-session dialogues, relative temporal expressions (e.g., ``last Friday'') often create ambiguity.  \textbf{TReMu}~\cite{ge2025tremu} employs a time-aware memorization mechanism to decouple the \textit{mention time} (session timestamp) from the inferred \textit{event time}. TReMu structures memory as ``timeline summaries,'' where events are grouped and indexed by their inferred absolute timesteps. This structure supports neuro-symbolic reasoning, where the agent generates Python code to perform precise arithmetic on dates before retrieving the corresponding timeline nodes.

\paragraph{Hierarchical Temporal Constraints}
\textbf{MemoTime}~\cite{tan2025memotime} is developed to prevent logical hallucinations in reasoning chains (e.g., retrieving an effect that occurred before its cause). This framework organizes the TKG reasoning process into a hierarchical ``Tree of Time.'' Unlike flat retrieval, this structure enforces \textit{temporal monotonicity}, ensuring that any retrieved reasoning path $e_1 \to e_2 \to e_3$ strictly adheres to chronological constraints ($t_1 \le t_2 \le t_3$). This operator-aware design allows the agent to prune semantically relevant but chronologically invalid evidence effectively.

\subsection{Hypergraph Structure}
\label{sec:hypergraph}
While binary graphs (connecting two entities) are efficient, they suffer from information loss when representing complex, multi-entity interactions (e.g., three drugs acting together to cause a side effect). Hypergraphs address this by employing hyperedges that can connect an arbitrary number of nodes, preserving the integrity of $n$-ary relations.

\paragraph{Information Integrity in N-ary Relations}
The primary motivation for hypergraph memory is to prevent the sparsity and semantic fragmentation caused by decomposing complex facts into binary edges. \textbf{HyperGraphRAG}~\cite{luo2025hypergraphrag} demonstrates that hypergraph-structured representation is information-theoretically more comprehensive than binary equivalents. By treating a natural language knowledge fragment and all its associated entities as a single hyperedge $e_i = (e_i^{text}, \{v_1, \dots, v_n\})$, the system enables ``dual-retrieval'', simultaneously retrieving relevant entities and the hyperedges that bind them, thereby accessing complete facts for generation.

\paragraph{Structural Dependency in Tabular Data}
Hypergraphs are particularly effective for structured data where relationships are inherently group-wise. \textbf{HyperG}~\cite{huang2025hyperg} models tabular knowledge using hypergraphs. It constructs distinct hyperedges for rows, columns, and the entire table to capture high-order dependencies, such as semantic consistency within columns and the hierarchical relationship between captions and cells. To enhance reasoning, HyperG utilizes a Prompt-Attentive Hypergraph Learning (PHL) module, which dynamically propagates attention between nodes and hyperedges based on the specific inquiry, effectively simulating a human-like focus on relevant data substructures.

\subsection{Hybrid Graph Architectures}
\label{sec:hybrid_graph}
Graph structures excel at precision and multi-hop reasoning but may lack the breadth of vector retrieval or the flexibility of unstructured buffers. Hybrid architectures fuse graphs with other data structures to balance these trade-offs, typically separating ``static knowledge'' from ``dynamic experience.''

\paragraph{Knowledge-Experience Decoupling}
A leading design pattern is to separate world rules from agent trajectories. \textit{Optimus-1}~\cite{li2024optimus} proposed a \textbf{Hybrid Multimodal Memory} for agent. It combines a \textit{Hierarchical Directed Knowledge Graph (HDKG)} to store static, structured game mechanics (modeled as directed acyclic graphs for crafting recipes) with an \textit{Abstracted Multimodal Experience Pool (AMEP)}. The AMEP functions as a dynamic vector store that retains multimodal success and failure trajectories. This separation allows the agent to ground planning in rigid graph-based knowledge while refining execution by retrieval-augmented experience from the pool.

\paragraph{External Graph with Internal Working Memory}
Another hybrid approach focuses on the interaction between an external massive graph and an internal lightweight state. The \textbf{KG-Agent} framework~\cite{jiang2025kg} integrates an external knowledge graph with an internal \textit{Knowledge Memory}. Unlike purely graph-based agents, KG-Agent maintains a structured scratchpad that iteratively updates the reasoning history, tool definitions, and intermediate observations retrieved from the external KG. This hybrid design enables the agent to navigate large-scale static graphs using a flexible, evolving internal context, bridging symbolic storage with neural reasoning.

In summary, different types of memory often require different approaches to graph construction, and there is no single paradigm that fits all scenarios. Knowledge memory, which is generally stable, structured, and context-independent, is well suited to graphs that emphasize relational or containment structure, such as knowledge graphs. These graphs can capture explicit relationships between factual entities, and enable efficient reasoning over globally valid information. In contrast, experience memory is dynamic, personalized, and context-specific, and often benefits from graphs that emphasize temporal sequences, interaction trajectories, or user–action networks. Such graphs capture the evolving patterns of agent interactions or preferences over time. The choice of graph type is therefore closely linked to the properties of the underlying memory. Factors such as sparsity, temporal dynamics, modality diversity, and the need for incremental updates all influence how nodes and edges are defined. Moreover, hybrid or multi-layered graphs are sometimes necessary to integrate both knowledge and experience memory. Overall, constructing graphs requires careful consideration of the characteristic of memory types.

\section{Memory Retrieval: Recalling the Past} \label{sec:retrieval}
\begin{figure*}[t]
    \centering
    \includegraphics[width=\linewidth]{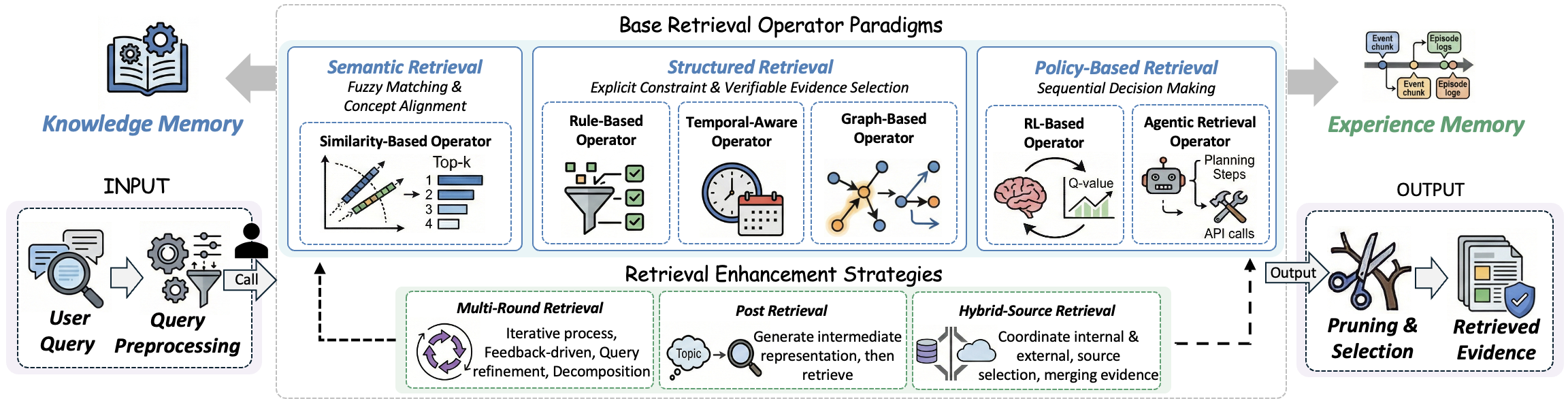}
    \caption{Retrieval pipeline architecture integrating base operators and enhancement 
strategies. \textbf{Left:} User queries go through preprocessing before retrieval. 
\textbf{Top:} Six base retrieval operators are organized into three paradigms: semantic, structured, and policy-based retrieval, which interact with knowledge and experience memory types. \textbf{Bottom:} Retrieval 
enhancement strategies layer on top of these operators: multi-round retrieval, post-retrieval, and hybrid-source 
retrieval coordinates internal memory with external resources. \textbf{Right:} Final 
pruning and selection produce ranked retrieved evidence for downstream reasoning with these operators and strategies.}
    \label{fig:retrieval}
\end{figure*}

These construction choices directly shape the memory usage. Once memory structures are fixed, the system must decide how to access them to support downstream reasoning. This motivates the retrieval stage, the next stage of the storage pipeline. The retrieval stage manipulates graph memory by defining executable operators. Base retrieval operators can be organized into three paradigms. These paradigms interact with the two memory roles: knowledge and experience.

\textbf{Semantic retrieval} including similarity-based operators operates over extracted text or multimodal chunks and their embeddings. It supports fuzzy matching and basic concept alignment. It is often used as an candidate generator for knowledge and experience memory.

\textbf{Structured retrieval} includes rule-based operators, temporal operators and graph-based operators. These operators execute explicit constraints over structured artifacts: knowledge graphs, hierarchies, temporal graphs, and hypergraphs. This enables verifiable and interpretable evidence selection: each retrieval decision is traceable 
through the underlying graph structure or rule set. Structured retrieval is particularly central to knowledge memory. For knowledge memory, correctness and consistency are primary concerns.

\textbf{Policy-based retrieval} includes RL and agentic retrieval operators that treat retrieval as sequential decision-making. The system selects which memory type to query and chooses which operators to apply. It allocates computational budgets and decides when to stop retrieval. Policy-based retrieval is especially important for experience memory, which is dynamic, personalized, and time-sensitive in nature.

In practice, systems commonly combine these basic operators, e.g., semantic anchoring $\rightarrow$ structured expansion $\rightarrow$ policy-controlled stopping and pruning.

\subsection{Categorization of Retrieval Techniques}
The retrieval process can be broken down into three basic operations: (i) query preprocessing, (ii) candidate retrieval, and (iii) pruning. They are usually executed as a simple pipeline that extracts a small set of relevant evidence from the memory graph. In this section, we introduce a set of base retrieval operators that can be flexibly composed to implement the latter two operations. Together with preprocessing, these operators form an end-to-end retrieval pipeline. Beyond these operators, we also describe retrieval enhancement strategies, which are auxiliary practices layered on top of the base operators to improve retrieval quality.

\subsubsection{Similarity-based Operator}

Similarity-based retrieval is a coarse retrieval operator to encode a user query into a vector and then retrieves the top-$k$ most similar memory entries in the embedding space~\cite{long2025seeing}. 

For knowledge memory, similarity-based operator primarily supports exact abstract concept or entity recall. For experience memory, the operator functions as specific memory unit recall, e.g., episodes, summaries, or agent states which match the current query. In practice, systems often combine similarity retrieval with experience memory and knowledge memory~\cite{chhikara2025mem0}. For experience memory, systems use summary graphs where the retriever first matches high-level summaries and then drills down to supporting raw corpus chunks. For knowledge memory, systems use knowledge graphs where retrieved entities or triples serve as anchors for subsequent graph expansion.

In simpler applications, direct similarity search can be effective~\cite{rasmussen2025zep}. It can also play a supporting role. For instance, it can use query-query~\cite{zhang2025g} similarity to filter unrelated memories. However, this approach has clear limitations:
\begin{itemize}[leftmargin=*, topsep=0pt, partopsep=0pt, parsep=0pt, itemsep=0pt]
    \item \textbf{Similarity does not guarantee relevance.} Lexically or semantically similar text may not match the specific memory needed for a task.
    \item \textbf{Poor multi-hop reasoning.} Answers to complex queries often depend on entities not in the original query. Pure similarity matching cannot bridge this gap.
    \item \textbf{Temporal awareness is often missing.} Memory retrieval in dynamic contexts requires understanding time and situational relevance, not just semantic proximity.
    \item \textbf{Scalability introduces noise.} As memory grows, the number of semantically similar segments increases. This leads to redundant or irrelevant retrievals, which degrade precision.
\end{itemize}
These challenges motivate more sophisticated retrieval beyond similarity search. Practical systems should leverage structured retrieval that align with the organization of memory. Moreover, retrieval should be policy-driven: an agent selects and composes operators conditioned on the underlying memory type.

\subsubsection{Rule-based Operator}
Rule-based operator in graph memory uses symbolic rules, and executable filters to decide relevant memories rather than  purely semantic similarity. 

For knowledge memory, rule-based operator serves two roles. It can act as a first-stage selector that preprocesses candidates for downstream retrieval, or as a post-retrieval validator that prunes retrieved memory by enforcing hard constraints. For experience memory, rule-based retrieval primarily supports temporal scoping over dynamic episodes such as dialogue histories and execution trajectories, since experiences are noisy and continually evolving~\cite{hou2024my}. In practice, for knowledge memory, rule-based operators can restrict candidates to compatible entity or relation types, prioritize high-confidence triples, and exclude triples flagged as conflicting by predefined rules. For experience memory, they leverage common rules such as time-window filters and task-phase constraints like  execution~\cite{salama2025meminsight}.

A common variant uses hand-crafted associative heuristics inspired by Hebbian-style updates~\cite{fisher2025neural}: memories that are frequently co-retrieved may have their links strengthened, and recently written or repeatedly referenced items can be up-weighted over time. Such lightweight update rules can improve coherence by retrieving bundles of mutually associated memories rather than isolated items~\cite{yu2025finmem}.

Rule-based retrieval also frequently applies deterministic symbolic filtering to satisfy explicit query constraints. In structured backends, e.g., SQL databases, these constraints can be compiled into executable queries or programs, e.g., SQL or Python code that queries and joins tables, improving precision and making the retrieval pipeline auditable~\cite{ge2025tremu}.

\subsubsection{Temporal-Based Operator}

Temporal-based operator is essential for handling queries that depend on when events occurred, tracking facts that change over time, and preserving the sequence of interactions in conversations~\cite{tan2025memotime,jonelagadda2025mnemosyne}.

For knowledge memory, general truths remain valid regardless of when they were learned, so temporal operators primarily serve as a filtering rule rather than as a core retrieval mechanism. For experience memory, which records interactions and events whose relevance shifts with time, temporal operators become primary. A user's interests evolve, recent tool failures matter more than old ones, and the system must distinguish between what worked last week versus last year. Temporal operators therefore rank episodes by how recently they occurred, downweight outdated preferences as time passes, and retrieve chains of events that form coherent narratives rather than isolated records from different time periods.

In practice, Zep~\cite{rasmussen2025zep} maintains explicit time windows for each fact, marking when it becomes valid and when it expires. This enables the system to determine whether a fact still applies at query time. LiCoMemory~\cite{huang2025licomemory} applies a decay function during ranking that reduces the weight of older facts while preserving highly relevant information. Systems also parse the user's query to infer the intended time range, such as extracting "last month" from "What did I discuss last month?", then limit retrieval to that window~\cite{zhang2025assomem}. These approaches improve recall for questions that inherently depend on timing.

\subsubsection{Graph-based Operator}
Graph-based retrieval operator traverses explicit relational links in experience or knowledge graph memories. It uses query-conditioned traversal to expand from anchor nodes into a task-relevant subgraph in the same graph or another graph.

For knowledge memory, a graph makes reasoning support explicit by encoding facts as linked units. Retrieval can enforce structured relational constraints and return justificatory structures such as short paths or induced subgraphs. It enables compositional queries that join multiple facts across the graph~\cite{jiang2025kg}. For experience memory, a graph supports reasoning by preserving the connectivity of episodes. It lets the agent reconstruct coherent state–action–reward chains and attribute rewards to specific conditions. It retrieves adjacent follow-up actions that are central to proactive adaptation~\cite{wu2025sgmem}.

In practice, knowledge and experience memory share a common working scenario: first, identify key entities as anchors; second, expand candidates via neighborhood traversal, e.g., within $n$ hops; third, score and prune nodes or paths to obtain compact evidence. In knowledge memory, anchors are typically entities or concepts, and traversal is often relation-constrained over a knowledge graph to support logic-like retrieval. In experience memory, anchors are commonly situation descriptors extracted from dialogue chunks and execution logs, such as tool outputs, or environment states, and traversal expands along event and temporal links to recover episodic context~\cite{cai2025simgrag}.

\paragraph{Intra-layer Traversal}
For instance, Mem0~\cite{chhikara2025mem0} follows an entity-centric approach that expands relations around identified entities. Zep~\cite{rasmussen2025zep} augments retrieval with breadth-first neighborhood expansion to collect candidate nodes or edges within a bounded hop radius, which can then be re-ranked by relevance. H-MEM~\cite{h-mem} uses index-based routing on a hierarchical memory tree to retrieve relevant content layer by layer. Some approaches further employ GNN-based models to obtain cross-layer representations for retrieval and scoring~\cite{wu2024can}.

\paragraph{Inter-layer Traversal}
In hierarchical graph memories, traversal also operates across abstraction levels via inter-layer edges, e.g., edges between summary  nodes and dialogue chunks, enabling bottom-up abstraction. G-Memory~\cite{zhang2025g} performs bi-directional traversal that synthesizes generalized strategies from an insight graph with detailed logs from an interaction graph. LiCoMemory~\cite{huang2025licomemory} uses explicit hyperlinks to traverse from abstract summaries to precise dialogue chunks containing supporting evidence. Trainable graph memory~\cite{xia2025experience} aggregates cross-layer path strengths across its query, transition path, and meta-cognition layers to derive relevance scores for retrieval.

\subsubsection{Reinforcement Learning-based Operator}
Reinforcement learning (RL) is increasingly used as a training operator to learn adaptive retrieval policies by optimizing downstream task rewards defined by the underlying memories. 

The two memory types serve different purposes: knowledge memory mainly provides stable grounding and constraint for this operator, whereas experience memory supports context-sensitive adaptation. Context-sensitive adaptation means adapting retrieval action to the agent's current context, such as the current task state, by recalling and ranking previously observed state-action-reward episodes that match the current state. They share the same purpose: to retrieve evidence that is consistently helpful while down-weighting evidence candidates that are semantically similar yet systematically misleading~\cite{wang2025mem}. This improves robustness beyond fixed similarity heuristics. In practice, RL-based retrieval operator can be applied to any structured memory systems whenever retrieval actions are well-defined, a reward signal reflecting retrieval utility is available, and the policy can be trained to balance performance gains against retrieval cost~\cite{xu2025memory}.

A common instantiation augments embedding-based candidate generation with a learned action-value function $Q(s,a)$, where the state $s$ summarizes the current query and optionally the agent state. The action $a$ corresponds to a retrieval decision such as selecting memory items, choosing a graph-navigation step, or issuing a tool call~\cite{wang2024crafting,zhou2025memento}

Beyond single-step selection, some works formulate retrieval as a sequential decision process and train a dedicated memory agent with policy optimization, e.g., PPO/GRPO~\cite{xia2025experience}. In a typical pipeline, the memory agent decides what to retrieve and in some designs also what to write into memory. It then passes the retrieved memory $M_{\text{ret}}$ to a possibly frozen answer agent to produce the final response. Rewards are computed from task-specific answer quality metrics, e.g., EM/F1 for QA, and used to update the retrieval policy to maximize  performance~\cite{yan2025memory}.

Since end-to-end online RL over LLMs can be expensive, an alternative is RL-free policy learning via retrospective labeling. In such schemes, an expert LLM evaluates which retrieved memories are genuinely useful, producing supervision signals to train a lightweight retriever, e.g., an MLP ranker. Another way is to use lightweight heuristic scorers to guide selection when the candidate set is large~\cite{yan2025memory,jia2025enhancing}. This offers a pragmatic trade-off between adaptivity and training cost when large-scale online RL is impractical~\cite{tan2025prospect}.

\subsubsection{Agent-based Operator}
Agent-based operator treats retrieval as an open planning-feedback loop where the agent can step beyond the boundaries of internal memory graphs~\cite{kim2024leveraging,anokhin2024arigraph}. The core distinction from RL-based approaches lies in the ability to call external tools and APIs to supplement internal memory, then store verified results back into memory for future use~\cite{rezazadeh2025collaborative}. This enables retrieval to extend beyond selecting from a fixed knowledge base.

For knowledge memory, the agent traverses the knowledge graph to gather supporting facts via self-planning~\cite{wang2025graphcogent}. Graph traversal allows the agent to discover chains of reasoning within the structured memory. When internal graph coverage is incomplete, the agent can leverage internal knowledge to justify information correctness. For experience memory, the agent does not traverse a graph but instead traces chains of actions across episodes to recover episodic context~\cite{sumers2023cognitive,hu2025hiagent}. Rather than navigating structured relationships, the agent reconstructs events from timestamped logs to make decisions~\cite{cai2025simgrag}.

The action space includes selecting the corresponding store and index and choosing between a knowledge-graph index for rules or facts and a temporal or hypergraph index for past episodes~\cite{tan2025cradle}. The agent navigates the selected memory structure by picking the next node or edge to visit or moving across abstraction layers. The agent maintains a bounded working memory that records the current question, the partial plan, which memory source each piece of evidence comes from, previously visited nodes, and the evidence collected so far. This enables closed-loop navigation decisions~\cite{li2024optimus,wang2025omni}.

Beyond pre-defined action spaces, the agent can write API calls, e.g., SQL queries or search requests, to retrieve information on-demand actively. This aligns retrieval decisions with the agent's planning process~\cite{yang2025llm,packer2023memgpt}. In complex systems, responsibilities can be split across multiple agents, e.g., candidate proposal vs. ranking, to improve robustness. 

\subsection{Retrieval Enhancement Strategies}
Beyond base retrieval operators, recent graph memory systems increasingly emphasize retrieval enhancement strategies. These methods do not define a new operator by themselves. Instead, they improve retrieval by adding extra steps around a base operator rather than one-shot look up. These strategies also help in both the two memory roles discussed earlier. Knowledge memory requires enhancement strategies that prioritize reliability and conflict handling, since stable facts and rules must remain consistent across retrievals. Experience memory requires enhancement strategies that prioritize temporal coherence, personalization, and iterative evidence gathering, since it captures what actually happened to the agent and user.

In this survey, we group them into three classes: i) \textbf{multi-round retrieval}, which increases search depth and coverage via repeated retrieval; ii) \textbf{post-retrieval}, which makes the query clearer by first generating an intermediate representation, e.g., topic or intent description and then retrieving; and iii) \textbf{hybrid-source retrieval}, which improves answer completeness by retrieving from both internal memory and external sources and then combining the results.

\subsubsection{Multi-round Retrieval}
Multi-round retrieval treats memory access as an iterative process rather than a single-pass query over the memory~\cite{yan2025general}. Each round generates the next retrieval query conditioned on the original query and previously retrieved memory, retrieves additional memories, and then aggregates and evaluates whether the accumulated evidence is sufficient~\cite{liu2025rcr,yuan2025memsearcher}. If sufficient, it terminates; otherwise, it triggers another round. More broadly, multi-round retrieval can be implemented with policies that decide whether to re-query, how to refine queries, and when to stop. This loop makes retrieval explicitly feedback-driven rather than static~\cite{yan2025general,zhang2025bridging}.

Moreover, some systems decompose a complex query into sub-queries~\cite{tan2025memotime} and perform retrieval at the sub-query level, optionally rewriting each sub-query to make the queries more expressive. This fine-grained decomposition can reduce semantic drift of query and enable targeted evidence gathering~\cite{zeng2024structural}.

For knowledge memory, extra rounds are often used to collect supporting facts or rules and to verify consistency across retrieved items. For experience memory, extra rounds are often used to reduce missing context, such as recent user requests, recent failures, or the latest environment changes. In practice, this also changes the stopping rule: knowledge-oriented loops stop when evidence is consistent and well supported, while experience-oriented loops stop when enough recent and relevant episodes have been gathered~\cite{han2025legomem,zhu2023ghost}.

\subsubsection{Post-retrieval}
While most memory-augmented systems retrieve memories before generation, post-retrieval follows a generate-then-retrieve pattern. The system first produces an intermediate representation, e.g., a topic, intent descriptor, hypothesized entities and relations, or a draft structure, and then retrieves based on that intermediate representation. For example, a model may generate a high-level topic prior to answering and use it to retrieve memories, or transform a query into an imagined subgraph and select candidate subgraphs that minimize graph distance~\cite{cai2025simgrag}. This design is motivated by a common failure mode in interactive settings: user queries are often ambiguous or poorly specified, and even LLM-based query rewriting may not reliably map such queries onto the right memories. By interposing a topic-generation step, post-retrieval is less sensitive to superficial query phrasing~\cite{wang2025mirix}.

Beyond explicit symbolic queries, there are also generative retrieval variants where the model uses its current latent reasoning state to generate a sequence of latent memory token representations. It then retrieves memories by embedding similarity in that latent space. The model can be trained using answer quality as the learning goal, encouraging latent memory token representations that surface more helpful memories~\cite{zhang2025memgen}.

For knowledge memory, the intermediate representation is often made closer to a canonical form so that the system can match rules and facts and then verify them. For experience memory, the intermediate representation often includes the missing context that the user did not say explicitly, such as the user goal, constraints, or what has already been tried. This is useful when the system must recover the right episode from many similar logs.

\subsubsection{Hybrid-source Retrieval}
Graph memory retrieval is increasingly studied in a hybrid-source setting, where the system coordinates internal memory with external resources. Here, external knowledge is treated as an additional retrievable resource alongside the memory store~\cite{yuan2025memsearcher}. Examples include a local document index, online search APIs that return titles and snippets with URLs, and task environments that can be accessed through an agent interface. A key challenge is source selection. The system must decide when to rely on internal memory versus external retrieval and how to merge evidence or resolve conflicts across sources.

Hybrid-source retrieval naturally supports knowledge memory and experience memory. External sources excel at providing information that changes frequently or extends beyond what the system has seen, which is closer to knowledge memory. Internal experience mainly provides personal and local details, such as users' previous purchase history or which tools failed in past attempts. When the two disagree, the merge rule should depend on what is being retrieved. If the system retrieves a fact, it should prioritize evidence that can be verified through multiple independent sources and traced back to authoritative origins, since outdated or unreliable external data is worse than trusted internal knowledge memory. If the system retrieves personal experience, it should prioritize internal records that match both the correct user and the correct time period, since external data cannot capture this individual~\cite{zhou2025mem1}.

\section{Memory Evolution: Learning Over Time} 
\label{sec:evolving}

\begin{figure*}[t]
    \centering
    \includegraphics[width=\linewidth]{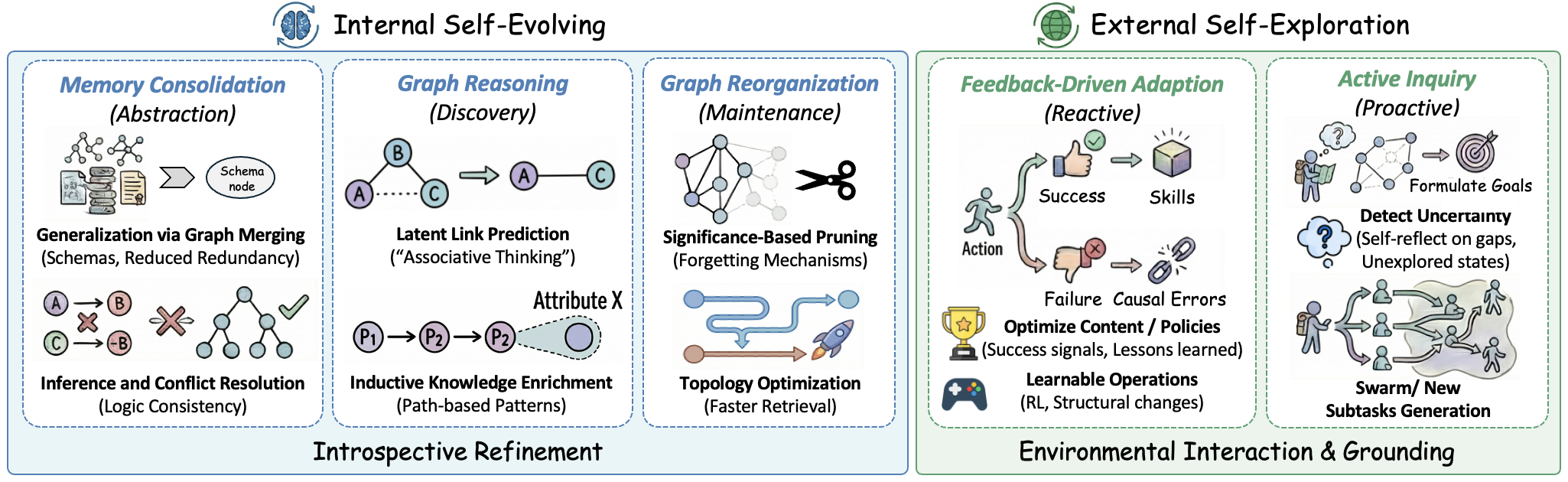}
    \caption{A taxonomy of Memory Evolution mechanisms. \textbf{(Left) Internal Self-Evolving} entails introspective refinement via memory consolidation, graph reasoning, and reorganization. \textbf{(Right) External Self-Exploration} involves grounding memory through environmental interaction, categorized into reactive feedback-driven adaptation and proactive active inquiry.}
    \label{fig:evolution}
\end{figure*}

As agent systems interact with dynamic environments over extended periods, their memory must not remain static but evolve to incorporate new information, resolve inconsistencies, and adapt to changing contexts. 
Graph-based memory structures are particularly well-suited for evolution due to their explicit modeling of relational connections, temporal dependencies, and validity that enable direct updates through node/edge/subgraph operations.
Drawing inspiration from cognitive science, where human memory consolidates and adapts via mechanisms like synaptic plasticity, graph-based agent memory evolves through internal self-reflection and external exploration. This dual approach addresses conflicts between new and old memories (e.g., outdated facts versus recent observations) while enhancing long-term coherence and adaptability.
As shown in Figure ~\ref{fig:evolution}, we categorize memory evolution into two complementary paradigms: i)\textbf{ internal self-evolving}, which focuses on intrinsic graph operations to maintain consistency without external input; ii)\textbf{ external self-exploration}, which leverages active interaction with the environment to ground and refine memory automatically. These mechanisms transform passive knowledge repositories into proactive, learning systems.

\subsection{Internal Self-Evolving}
Internal self-evolving treats the agent's memory graph as a closed-loop system capable of introspective refinement. This process is analogous to human memory consolidation during sleep or rest, where the brain organizes recent experiences, abstracts general rules, and forgets trivial details without requiring new external sensory input. In the context of graph-based agent memory, this involves reorganizing the graph topology to enhance retrieval efficiency, logical consistency, and generalization capabilities. Unlike traditional storage that simply appends the logs, internal evolution applies structural transformations to the graph, typically focusing on three key aspects: \textbf{Memory Consolidation}, \textbf{Graph Reasoning}  and \textbf{Graph Reorganization}.

\subsubsection{Memory Consolidation}
The core of internal evolution is the abstraction of high-level knowledge from raw experiential data. Agents accumulate vast amounts of experience memory (specific trajectories or interaction logs). Internal evolution mechanisms~\cite{rasmussen2025zep,nan2025nemori,wang2025mem} analyze these episodes to induce generalized knowledge.
\begin{itemize}
\item \textbf{Generalization via Graph Merging:} When an agent observes multiple similar distinct events, it can merge these subgraph instances into a generalized schema node~\cite{edge2024local}. This reduces storage redundancy and creates a canonical representation of "skills" or "facts"~\cite{kynoch2023recallmadaptablememorymechanism,tang2025agent}. More advanced methods like Mem0~\cite{chhikara2025mem0} and FLEX~\cite{cai2025flex} employ LLM-based semantic gating, where the model evaluates the information gain of new trajectories
\item \textbf{Inference and Conflict Resolution:} Through internal reasoning (e.g., logical inference or graph traversal), the agent can detect implicit contradictions between nodes. For instance, if node $A$ implies $B$, but node $C$ implies $\neg B$, the system triggers a self-correction process to resolve the conflict based on confidence scores or temporal recency, updating the graph structure to maintain logical coherence~\cite{Shinn2023ReflexionAA,sun2024thinkongraph,RRP}.
\end{itemize}

\subsubsection{Graph Reasoning}
Beyond consolidating existing nodes, internal evolution also entails discovering latent connections to address memory sparsity. Agents actively scan their memory graphs to identify missing edges or infer new relationships that were not explicitly observed but can be deduced from existing facts. This process transforms the memory from a sparse set of isolated trajectories into a densely connected knowledge web.
\begin{itemize}
\item \textbf{Latent Link Prediction:} Utilizing the semantic capabilities of LLMs, agents can predict potential relationships between disjoint subgraphs. For instance, if the memory contains $(A \xrightarrow{cause} B)$ and $(B \xrightarrow{cause} C)$, the agent can autonomously infer and insert a transitive edge $(A \xrightarrow{leads\_to} C)$~\cite{sun2024thinkongraph,luo2024reasoning}. This mechanism mimics the ``associative thinking" in human cognition, allowing agents to ``connect the dots" between temporally distant events without new external input.
\item \textbf{Inductive Knowledge Enrichment:} Agents can employ path-based reasoning to derive new attributes or facts. By traversing the graph structure, the system identifies patterns and enriches the graph with inferred ``covariates" or high-order relationships~\cite{edge2024local}. This effectively mitigates the incompleteness of the initial memory construction, ensuring that future retrieval can access logically derivative information that was never explicitly written into the logs.
\end{itemize}

\subsubsection{Graph Reorganization}
Infinite accumulation of memory leads to retrieval latency and noise. Internal self-evolving implements ``forgetting" mechanisms akin to biological systems to maintain the graph's health and quality. \begin{itemize} 
\item \textbf{Significance-Based Pruning:} Agents employ algorithms (e.g., PageRank variants or decay functions) to evaluate the utility of memory nodes. Nodes that are rarely accessed or have low contribution to recent decision-making are pruned or compressed~\cite{zhong2024memorybank,kang2025memory,packer2023memgpt}. 
\item \textbf{Topology Optimization:} This involves restructuring the edges to shorten the path length between frequently associated concepts~\cite{kynoch2023recallmadaptablememorymechanism,hippo2}. By increasing the edge weights or creating shortcuts between correlated nodes, the agent optimizes the graph for faster future retrieval, compiling its experience into a more efficient structure. 
\end{itemize}

\subsection{External Self-Exploration}

While internal evolution refines memory based on intrinsic consistency, it cannot verify the \textit{validity} of knowledge against the real world. \textbf{External Self-Exploration} bridges this gap by grounding memory evolution in environmental interaction. Instead of passively recording all events, effective exploration uses environmental feedback (e.g., success signals, errors) to distinguish signal from noise and proactively seeks missing information. We categorize this broadly into feedback-driven adaptation (reactive) and active inquiry (proactive).

The first paradigm, \textbf{feedback-driven adaptation}, focuses on optimizing memory content and management policies based on the outcomes of past actions. In open-ended environments, raw interaction logs are often noisy; therefore, agents utilize task execution results as supervisory signals to distill actionable knowledge. Methods like ExpeL~\cite{zhao2024expel} and Matrix~\cite{xu2024matrix} differentiate processing based on success or failure: successful trajectories are crystallized into reusable skills, while failed ones undergo comparative analysis to isolate causal errors, explicitly encoding ``lessons learned'' to favor high-utility behaviors. This evolutionary logic extends beyond content to the memory management policy itself. Rather than relying on fixed rules, systems like Memory-R1~\cite{yan2025memory} and Inside Out~\cite{gekhman2025inside} treat memory operations (such as addition or deletion) as learnable actions within a Reinforcement Learning framework. By receiving rewards for downstream task accuracy, the agent autonomously learns an optimal policy for maintaining the graph. Similarly, MemEvolve~\cite{zhang2025memevolve} applies this feedback-driven principle to the architectural level, dynamically adjusting storage structures and indexing mechanisms to tailor the system to the changing complexity of the deployment environment.

However, relying solely on reactive adaptation suffers from coverage bias, where the agent's knowledge remains limited to the specific tasks it has been assigned. To overcome this, the second paradigm, \textbf{active inquiry}, transitions the agent from a passive learner to a proactive explorer. Advanced frameworks empower agents to detect uncertainty in their current graph and autonomously formulate goals to resolve it. For instance, ProMem~\cite{yang2026beyond} enables agents to critically self-reflect on missing nodes or ambiguous edges, subsequently generating specific queries to fill these knowledge gaps. AgentEvolver~\cite{zhai2025agentevolver} expands this capability to the task space by autonomously generating new sub-tasks that force navigation through unexplored states, preemptively populating memory with diverse experiences. To further accelerate this process, KIMI K2.5~\cite{moonshot2026k25} introduces a scalable swarm mechanism, spawning parallel sub-agents to explore disjoint branches of a problem space. This allows the central memory to rapidly ingest diverse perspectives and edge cases, transforming the system from a passive container of experience into an active constructor of knowledge.


Despite progress in feedback-driven adaptation and active inquiry, explicitly leveraging graph structure for exploration remains underexplored. Here, we give some suggestions about using the graph in self-exploration. First, topology-guided exploration can prioritize sparsely-connected clusters or bridge disconnected subgraphs to systematically improve knowledge coverage and connectivity. Then, multi-granularity strategies coordinate discovery across seeking high-level relational patterns while populating concrete instances through targeted interactions. These approaches transform memory graphs from passive repositories into active architectures that dynamically construct and refine their own knowledge boundaries.

\section{Open-sourced Libraries and Benchmarks} \label{sec:libraries_and_benchmarks}
\subsection{Open-sourced Libraries}
In Table~\ref{tab:library_checklist} in Appendix~\ref{app:open-sourced libraries}, we provide a systematic comparison of eleven representative open-source memory libraries across key functional dimensions. 

Several libraries offer graph-based memory representation and utilization, including Cognee~\cite{markovic2025optimizing}, Mem0~\cite{chhikara2025mem0}, OpenMemory, MemMachine, Memary, and Graphiti. 
Graph-based structures naturally couple with structured retrieval, enabling multi-hop or relational queries essential for reasoning over entities, events, or concepts. In terms of functional coverage, OpenMemory and Mem0~\cite{chhikara2025mem0} act as the most comprehensive graph memory tools, supporting memory construction, interaction-driven updates, lifecycle management, temporal awareness, and graph management. Graph-based memory tools can be directly integrated into agent architectures to dynamically leverage structured knowledge, supporting long-term retrieval, temporal reasoning, and multi-step task execution. Cognee~\cite{markovic2025optimizing} provides queryable graph embeddings, Mem0~\cite{chhikara2025mem0} and OpenMemory support session-aware memory updates, and Graphiti enables temporal graph reasoning for multi-step planning.
For non-graph memory systems, memory construction is predominantly interaction or session-driven, such as LangMem, LightMem~\cite{fang2025lightmem}, and O-Mem~\cite{wang2025omni}, with mechanisms to update memory incrementally and support retrieval. While they provide retrieval functionalities fundamentally based on embedding similarity. Although lightweight tools like Memori and MemMachine focus on modular memory management, prioritizing ease of integration and supporting agent conditioning through APIs.

\subsection{Datasets and Benchmarks}



\begin{table*}[htbp]
\centering
\scriptsize
\setlength{\tabcolsep}{1.6pt}
\renewcommand{\arraystretch}{1.00}
\caption{Agent memory benchmarks grouped by scenario.}
\label{tab:agent-memory-benchmarks}
\resizebox{\textwidth}{!}{
\begin{tabular}{
@{} >{\raggedright\arraybackslash}p{0.14\textwidth}  
   >{\raggedright\arraybackslash}p{0.1\textwidth}  
   >{\raggedright\arraybackslash}p{0.12\textwidth}  
   >{\raggedright\arraybackslash}p{0.27\textwidth}  
   >{\centering\arraybackslash}p{0.1\textwidth}    
   >{\raggedright\arraybackslash}p{0.15\textwidth}  
   >{\raggedright\arraybackslash}p{0.120\textwidth}  
@{}}
\toprule
\textbf{Name} & \textbf{Scenario} & \textbf{Modality} & \textbf{Feature} & \textbf{Environment} & \textbf{Memory type} & \textbf{Link(Feb 2026)} \\
\midrule

LoCoMo~\cite{maharana2024evaluating} & Interaction & Text+Image & Long conversational memory & real & Factual &
\href{https://snap-research.github.io/locomo/}{\textbullet\ Website} \\

LongMemEval~\cite{wu2025longmemeval} & Interaction & Text & Long-term interactive memory & simulated & Factual &
\href{https://github.com/xiaowu0162/LongMemEval}{\textbullet\ GitHub} \\

MemoryAgentBench~\cite{hu2025evaluating} & Interaction & Text & Multi-turn interactions & simulated & Factual + Experiential &
\href{https://github.com/HUST-AI-HYZ/MemoryAgentBench}{\textbullet\ GitHub} \\

MEMTRACK~\cite{deshpande2025memtrack} & Interaction & Text+Code+Logs & Long-term interactive memory & simulated & Factual + Experiential &
\href{https://arxiv.org/abs/2510.01353}{\textbullet\ Website} \\

MADial-Bench~\cite{he2025madial} & Interaction & Text & Memory-augmented dialogue generation & simulated & Factual &
\href{https://github.com/hejunqing/MADial-Bench}{\textbullet\ GitHub} \\

MemSim~\cite{zhang2024memsim} & Interaction & Text & Bayesian memory simulation & simulated & Factual + Experiential &
\href{https://github.com/nuster1128/MemSim}{\textbullet\ GitHub} \\

ChMapData~\cite{wu2025interpersonal} & Interaction & Text & Memory-aware proactive dialogue & simulated & Factual &
\href{https://github.com/FrontierLabs/MapDia}{\textbullet\ GitHub} \\

MSC~\cite{xu2022beyond} & Interaction & Text & Multi-session chat & simulated & Factual &
\href{https://parl.ai/projects/msc/}{\textbullet\ Website} \\

MMRC~\cite{xue2025mmrc} & Interaction & Text+Image & Multi-modal real-world conversation & simulated & Factual &
\href{https://github.com/haochen-MBZUAI/MMRC}{\textbullet\ GitHub} \\

MemBench~\cite{tan2025membench} & Interaction & Text & Interactive scenarios & simulated & Factual + Experiential &
\href{https://github.com/import-myself/Membench}{\textbullet\ GitHub} \\   

StoryBench~\cite{wan2025storybench} & Interaction & Text & Interactive fiction memory & mixed & Factual + Experiential &
\href{https://arxiv.org/abs/2506.13356}{\textbullet\ Website} \\ 

DialSim~\cite{kim2024dialsim} & Interaction & Text & Multi-dialogue understanding & real & Factual + Experiential &
\href{https://arxiv.org/abs/2406.13144}{\textbullet\ Website} \\ 
RealMem~\cite{bian2026realmem} & Interaction & Text &
Project-oriented long-term memory interaction & simulated &
Factual + Experiential &
\href{https://github.com/AvatarMemory/RealMemBench}{\textbullet\ GitHub} \\

\midrule

PersonaMem~\cite{jiang2025know} & Personalization & Text & Dynamic user profiling & simulated & Factual &
\href{https://github.com/bowen-upenn/PersonaMem}{\textbullet\ GitHub} \\ 

PerLTQA~\cite{du2024perltqa} & Personalization & Text & Social personalized interactions & simulated & Factual &
\href{https://arxiv.org/abs/2402.16288}{\textbullet\ Website} \\ 

MemoryBank~\cite{zhong2024memorybank} & Personalization & Text & User memory updating & simulated & Factual &
\href{https://github.com/zhongwanjun/MemoryBank-SiliconFriend}{\textbullet\ GitHub} \\  

MPR~\cite{zhang2025explicit} & Personalization & Text & User personalization & simulated & Factual &
\href{https://github.com/nuster1128/MPR}{\textbullet\ GitHub} \\  

PrefEval~\cite{zhao2025llms} & Personalization & Text & Personal preferences & simulated & Factual &
\href{https://prefeval.github.io/}{\textbullet\ Website} \\ 

LOCCO~\cite{jia2025evaluating} & Personalization & Text & Chronological conversations & simulated & Factual &
\href{https://github.com/JamesLLMs/LoCoGen}{\textbullet\ GitHub} \\    

\midrule

WebChoreArena~\cite{miyai2025webchorearena} & Web & Text+Image & Tedious web browsing & real & Factual + Experiential &
\href{https://github.com/WebChoreArena/WebChoreArena}{\textbullet\ GitHub} \\    

MT-Mind2Web~\cite{deng2024multi} & Web & Text & Conversational web navigation & real & Factual + Experiential &
\href{https://github.com/magicgh/self-map}{\textbullet\ GitHub} \\ 

WebShop~\cite{yao2022webshop} & Web & Text+Image & E-commerce web interaction & simulated & Experiential &
\href{https://github.com/princeton-nlp/WebShop}{\textbullet\ GitHub} \\  

WebArena~\cite{zhou2024webarena} & Web & Text+Image & Web interaction & real & Experiential &
\href{https://github.com/web-arena-x/webarena}{\textbullet\ GitHub} \\  

MMInA~\cite{tian2025mmina} & Web & Text+Image & Multihop web agent & real & Factual + Experiential &
\href{https://mmina.cliangyu.com/}{\textbullet\ Website} \\

\midrule

NQ~\cite{kwiatkowski2019natural} & LongContext & Text & Natural question answering & simulated & Factual &
\href{https://ai.google.com/research/NaturalQuestions}{\textbullet\ Website} \\

TriviaQA~\cite{joshi2017triviaqa} & LongContext & Text & Large-scale question answering & simulated & Factual &
\href{http://nlp.cs.washington.edu/triviaqa/}{\textbullet\ Website} \\

PopQA~\cite{mallen2023not} & LongContext & Text & Adaptive retrieval augmentation & simulated & Factual &
\href{https://github.com/AlexTMallen/adaptive-retrieval}{\textbullet\ GitHub} \\

HotpotQA~\cite{yang2018hotpotqa} & LongContext & Text & Explainable multi-hop QA & simulated & Factual &
\href{https://hotpotqa.github.io/}{\textbullet\ Website} \\

2wikimultihopQA~\cite{ho2020constructing} & LongContext & Text & Multi-hop QA & simulated & Factual &
\href{https://github.com/Alab-NII/2wikimultihop}{\textbullet\ GitHub} \\

Musique~\cite{trivedi2022musique} & LongContext & Text & Multi-hop QA & simulated & Factual &
\href{https://github.com/stonybrooknlp/musique}{\textbullet\ GitHub} \\

LongBench~\cite{bai2024longbench} & LongContext & Text & Long-context understanding & mixed & Factual &
\href{ https://github.com/THUDM/LongBench}{\textbullet\ GitHub} \\

LongBench v2~\cite{bai2025longbench} & LongContext & Text & Long-context multitasks & mixed & Factual &
\href{ https://github.com/THUDM/LongBench}{\textbullet\ GitHub} \\

RULER~\cite{hsieh2024ruler} & LongContext & Text & Long-context retrieval & simulated & Factual &
\href{https://github.com/NVIDIA/RULER}{\textbullet\ GitHub} \\

BABILong~\cite{kuratov2024babilong} & LongContext & Text & Long-context reasoning & simulated & Factual &
\href{https://github.com/booydar/babilong}{\textbullet\ GitHub} \\

MM-Needle~\cite{wang2025multimodal} & LongContext & Text+Image & Multimodal needle retrieval & simulated & Factual &
\href{https://mmneedle.github.io/}{\textbullet\ Website} \\

HaluMem~\cite{chen2025halumem} & LongContext & Text & Memory hallucination eval & simulated & Factual &
\href{https://github.com/MemTensor/HaluMem}{\textbullet\ GitHub} \\

\midrule

MemoryBench~\cite{ai2025memorybench} & Continual & Text & Continual learning & simulated & Factual + Experiential &
\href{https://github.com/LittleDinoC/MemoryBench}{\textbullet\ GitHub} \\

LifelongAgentBench~\cite{zheng2025lifelongagentbench} & Continual & Text & Lifelong learning & simulated & Factual + Experiential &
\href{https://arxiv.org/abs/2505.11942}{\textbullet\ Website} \\

StreamBench~\cite{wu2024streambench} & Continual & Text & Continuous online learning & simulated & Factual + Experiential &
\href{https://stream-bench.github.io/}{\textbullet\ Website} \\

Evo-Memory~\cite{wei2025evo} & Continual & Text & Test-time learning & simulated & Factual + Experiential &
\href{https://arxiv.org/abs/2511.20857}{\textbullet\ Website} \\

\midrule

Ego4D~\cite{grauman2022ego4d} & Environments & Video+Audio & Egocentric episodic memory & real & Experiential &
\href{https://ego4d-data.org/}{\textbullet\ Website} \\

EgoLife~\cite{yang2025egolife} & Environments & Video+Audio & Long-context life QA & real & Experiential &
\href{https://egolife-ai.github.io/}{\textbullet\ Website} \\

ALFWorld~\cite{shridhar2020alfworld} & Environments & Text & Household tasks & simulated & Factual + Experiential &
\href{https://alfworld.github.io/}{\textbullet\ Website} \\

BabyAI~\cite{chevalier2018babyai} & Environments & Text & Language navigation & simulated & Experiential &
\href{https://arxiv.org/abs/1810.08272}{\textbullet\ Website} \\

ScienceWorld~\cite{wang2022scienceworld} & Environments & Text & Multi-step science experiments & simulated & Factual + Experiential &
\href{https://github.com/allenai/ScienceWorld}{\textbullet\ GitHub} \\

AgentGym~\cite{xi2025agentgym} & Environments & Text & Multiple environments & mixed & Experiential &
\href{https://agentgym.github.io/}{\textbullet\ Website} \\

AgentBoard~\cite{chang2024agentboard} & Environments & Text & Multi-round interaction & mixed & Experiential &
\href{https://github.com/hkust-nlp/AgentBoard}{\textbullet\ GitHub} \\

\midrule

SWE-Bench~\cite{jimenez2023swe} & Tool/Gen & Text+Code & Code repair & real & Experiential &
\href{https://www.swebench.com/}{\textbullet\ Website} \\ 

GAIA~\cite{mialon2023gaiabenchmarkgeneralai} & Tool/Gen & Text & Deep research tasks & real & Experiential &
\href{https://huggingface.co/gaia-benchmark}{\textbullet\ Website} \\ 

xBench-DS~\cite{chen2025xbenchtrackingagentsproductivity} & Tool/Gen & Text+Image & Deep-search evaluation & real & Experiential &
\href{https://xbench.org/}{\textbullet\ Website} \\ 

ToolBench~\cite{qin2023toolllmfacilitatinglargelanguage} & Tool/Gen & Text$\rightarrow$API & API tool use & real & Experiential &
\href{https://github.com/OpenBMB/ToolBench}{\textbullet\ Website} \\ 

GenAI-Bench~\cite{li2024genaibenchevaluatingimprovingcompositional} & Tool/Gen & Text+Image & Visual generation eval & real & Experiential &
\href{https://arxiv.org/abs/2406.13743}{\textbullet\ Website} \\

\bottomrule
\end{tabular}}
\end{table*}

Unlike standard NLP tasks, evaluating \emph{agent memory} should account for information dispersed throughout extended interactions and evolving system environments. Effective benchmarks in this area prioritize an agent’s ability to reuse observed data despite the limitations of finite context windows and computational costs. In the following, we highlight key benchmarks tailored for memory-augmented agents. These selections, summarized in Table~\ref{tab:agent-memory-benchmarks}, are evaluated against a unified set of criteria including modality, environment realism, and memory type to provide a comparative benchmark overview.

\paragraph{Scenario taxonomy}


We categorize existing benchmarks through a scenario-based taxonomy that reflects the diverse application settings of memory-augmented agents. This classification is based on three key aspects: interaction patterns ranging from multi-turn dialogues to long-horizon tasks, working interfaces including web-based or embodied and tool-assisted environments, and the time spans of information reuse. Following this structure, we identify seven representative scenarios: Interaction, Personalization, Web, LongContext, Continual, Environments, and Tool/Gen. 


\subsubsection{Interaction: Multi-turn and Cross-session Conversational Memory}

Benchmarks in the Interaction scenario focus on an agent's ability to maintain continuity across multi-turn and cross-session dialogues. In these settings, relevant information introduced early in the conversation must be accurately recalled and applied later. This category includes datasets such as LoCoMo~\cite{maharana2024evaluating}, LongMemEval~\cite{wu2025longmemeval}, MemoryAgentBench~\cite{hu2025evaluating}, MEMTRACK~\cite{deshpande2025memtrack}, MADial-Bench~\cite{he2025madial}, MemSim~\cite{zhang2024memsim}, ChMapData~\cite{wu2025interpersonal}, MSC~\cite{xu2022beyond}, MMRC~\cite{xue2025mmrc}, MemBench~\cite{tan2025membench}, StoryBench~\cite{wan2025storybench}, DialSim~\cite{kim2024dialsim}, and RealMem~\cite{bian2026realmem}. These benchmarks prioritize long-range context, consistency across different sessions, and the retrieval of previously mentioned user-provided facts during an ongoing dialogue.

The evaluation typically centers on memory recall over extended histories, contextual reuse of information to produce coherent responses, and consistency maintenance to avoid self-contradiction. These capabilities are fundamental for assistant-style agents where memory failures directly lead to a degraded user experience. Common metrics include task-level accuracy, retrieval-oriented measures like Recall@k, and dialogue-level consistency rates. Furthermore, some benchmarks track success rates over multi-turn tasks to see if recalled information is effectively integrated into responses. A notable limitation is that many Interaction benchmarks lack explicit supervision for memory updates when dealing with conflicting facts. Consequently, the mechanism by which agents overwrite or forget outdated information is less systematically evaluated than their ability to recall it.

\subsubsection{Personalization: User Profiling, Preferences, and Memory Updates}

Personalization benchmarks examine an agent's ability to manage persistent user-centric facts and profile attributes, as seen in PersonaMem~\cite{jiang2025know}, PerLTQA~\cite{du2024perltqa}, MemoryBank~\cite{zhong2024memorybank}, MPR~\cite{zhang2025explicit}, PrefEval~\cite{zhao2025llms}, and LOCCO~\cite{jia2025evaluating}. These benchmarks evaluate whether an agent can build a stable user model and integrate new user information over time. For real-world assistants, a primary challenge is avoiding persona drift where the agent forgets or contradicts previously established preferences. However, a current limitation of these tasks is their frequent reliance on clear supervision regarding what should be stored. In contrast, practical deployment necessitates selective writing and privacy-aware retention, both of which remain less addressed in existing efforts.

\subsubsection{Web: Long-horizon Browsing and Multi-step Online Tasks}

Web benchmarks evaluate memory within extended action trajectories where agents must track environmental states and intermediate results across numerous steps. Datasets like WebShop~\cite{yao2022webshop} and WebArena~\cite{zhou2024webarena} focus on e-commerce and functional website interactions. WebChoreArena~\cite{miyai2025webchorearena} targets complex browsing routines, while MT-Mind2Web~\cite{deng2024multi} emphasizes conversational navigation and MMInA~\cite{tian2025mmina} assesses multi-hop web interactions. These tasks place heavy demands on experiential memory because agents must often cache page states to prevent redundant actions. They also highlight the necessity of resource-efficient memory given that excessive tool calls can lead to high operational costs. A prevailing challenge is that success in these benchmarks can sometimes be achieved via simple heuristics, meaning that isolating the specific impact of memory requires controlled settings such as limiting memory capacity.

\subsubsection{LongContext: Long-document Understanding and Retrieval}

LongContext benchmarks measure agent performance under high-volume inputs and retrieval-intensive settings. Established QA suites including NQ~\cite{kwiatkowski2019natural}, TriviaQA~\cite{joshi2017triviaqa}, PopQA~\cite{mallen2023not}, HotpotQA~\cite{yang2018hotpotqa}, 2wikimultihopQA~\cite{ho2020constructing}, and Musique~\cite{trivedi2022musique} test evidence aggregation and multi-step reasoning. More recent frameworks like LongBench~\cite{bai2024longbench} and LongBench v2~\cite{bai2025longbench} offer multi-task evaluations, while RULER~\cite{hsieh2024ruler}, BABILong~\cite{kuratov2024babilong}, MM-Needle~\cite{wang2025multimodal}, and HaluMem~\cite{chen2025halumem} focus on needle-in-a-haystack retrieval and hallucination evaluation. While these benchmarks are essential for modeling evidence access, they are not always perfect indicators of agent memory. Many of these tasks remain single-turn and do not require the agent to actively write to a persistent memory store, potentially conflating long-context processing with dedicated memory mechanisms.

\subsubsection{Continual: Lifelong Learning and Test-time Adaptation}

Continual benchmarks assess whether agents can improve over time without experiencing catastrophic forgetting, typically under streaming or sequential task distributions. Frameworks such as MemoryBench~\cite{ai2025memorybench}, LifelongAgentBench~\cite{zheng2025lifelongagentbench}, StreamBench~\cite{wu2024streambench}, and Evo-Memory~\cite{wei2025evo} capture elements of online updates and test-time adaptation. This category represents lifelong memory in its strictest sense and requires models to maintain proficiency in earlier tasks while acquiring new knowledge. Despite its importance, this area lacks standardized reporting because metrics for forgetting and transfer gains vary significantly. Furthermore, it is often difficult to discern whether performance gains stem from parametric updates or retrieval over past logs.

\subsubsection{Environments: Embodied and Interactive Worlds}

Environment-based benchmarks evaluate agents in simulated or physical interactive settings where memory must distill observations under partial observability. Ego4D~\cite{grauman2022ego4d} and EgoLife~\cite{yang2025egolife} focus on egocentric episodic memory and multimodal life-logging. ALFWorld~\cite{shridhar2020alfworld} and BabyAI~\cite{chevalier2018babyai} emphasize instruction following and navigation, while ScienceWorld~\cite{wang2022scienceworld} tests multi-step experimentation. Broader suites like AgentGym~\cite{xi2025agentgym} and AgentBoard~\cite{chang2024agentboard} offer multi-round evaluations using planning-centric analysis. These benchmarks primarily test experiential memory and robustness across environmental variations. However, since performance is often tied to environment-specific skills, claiming memory-related benefits requires carefully controlling for planning and tool-use variables.

\subsubsection{Tool/Gen: Tool Use and Workflow Execution}

Tool/Gen benchmarks evaluate memory within workflows involving external APIs and iterative reasoning. ToolBench~\cite{qin2023toolllmfacilitatinglargelanguage} focuses on API invocation, while SWE-Bench~\cite{jimenez2023swe} targets software engineering through iterative debugging. GAIA~\cite{mialon2023gaiabenchmarkgeneralai}, xBench-DS~\cite{chen2025xbenchtrackingagentsproductivity}, and GenAI-Bench~\cite{li2024genaibenchevaluatingimprovingcompositional} measure complex research and generation behaviors. These tasks emphasize process memory or the ability to retain intermediate hypotheses and failed attempts. They also underscore operational issues, e.g., traceability and the financial cost of retries. A significant hurdle is evaluation complexity because success hinges on environment stability and the design of specialized scoring scripts.

\paragraph{Synthesis of Evaluation Landscapes}
Overall, these benchmarks provide a broad perspective on agent capabilities by testing them in interactive and long-horizon tasks. While memory is not always the only metric, success in these environments depends on the ability of an agent to store relevant data and use past experience for current decisions. Future studies can improve these evaluations by measuring the specific impact of memory through ablation tests and focusing on how agents handle changing information in dynamic settings. By using clear efficiency metrics and ensuring that environments are reproducible, these benchmarks help provide a better understanding of memory behaviors in practice.

\paragraph{Strategic Benchmark Selection}
Choosing the right benchmark depends on which memory capability is being studied. Interaction and Personalization datasets are best for testing conversational persistence. Web and Environments are more suitable for evaluating how agents manage long sequences of actions and experiential data. LongContext tasks remain the standard for checking fact retrieval in large inputs. For research on agents that learn over time, Continual benchmarks show how well information is kept, while Tool/Gen tasks evaluate memory during the execution of complex technical steps.

\section{Applications} 
\label{sec:applications}

The applicability of (graph-based) agent memory ranges from conversational chatbots to embodied robots and science agents. By addressing challenges, including long-term knowledge retention, personalized interaction, multi-step reasoning, and self-evolution, memory can enhance the effectiveness and reliability of LLM agents in a wide range of application domains. This section systematically discusses both current and prospective applications.

\subsection{Conversational Agents}
Conversational agents, such as Claude\footnote{\url{https://claude.ai/}} and ChatGPT\footnote{\url{https://chatgpt.com/}}, are one of the most common use cases of LLM-based systems. The agents face challenges in maintaining coherent and personalized dialogues within a context for a long period of time in a multi-session dialogue and require sophisticated memory systems to effectively update their knowledge and user preferences. 

Early studies focused on the controllability and stability of memory systems. The Memory Sandbox~\cite{huang2023memory} and LD-Agent~\cite{li2025hello} initiatives laid the foundation by highlighting the transparency of memory systems and the distinction between event memory and personality to enhance the credibility of the memory system. Contemporary studies on memory moved on to tackle the issue of context fragmentation in multi-session dialogue systems by using structural optimization methods. SeCom~\cite{pan2025secom} optimizes the dialogue structure by extracting topics in dialogue systems, while SGMem~\cite{wu2025sgmem} relies on semantic graphs to connect fragmented dialogue sessions. These studies upgrade memory systems from simple linear structures to knowledge graphs enabling more precise memory recall.

Aside from memory storage, advanced memory-based dialogue agents need to be equipped with dynamic reasoning and time evolution capabilities. RMM~\cite{tan2025prospect} extends memory systems with reflection-based self-correction mechanisms, whereas TReMu~\cite{ge2025tremu} uses temporal knowledge graphs to capture intricate time-based relationships. Nevertheless, the trend of increasing memory system complexity is challenged by the recent ENGRAM~\cite{patel2025engram} project, which showed that an efficient memory system with a basic typed memory structure consisting of episodic, semantic, and procedural memory types can achieve comparable performance to complex memory systems.

\subsection{Code Agents}
The software engineering processes for code generation~\cite{dong2025survey} and code simulation~\cite{islam2025codesim} pose a unique challenge for the memory component of the agent, as they require following rigid structural requirements and logical flow of software programming. In the initial stages, the problem of task confusion was addressed in both MetaGPT~\cite{hong2023metagpt} and ChatDev~\cite{qian2024chatdev} by modeling conventional human workflows and definitions of roles and responsibilities. SWE-agent~\cite{yang2024swe} further enhanced this approach with the inclusion of Agent-Computer Interfaces, which allow the agent to perform operations on source control systems. However, as the task becomes more intricate, linear approaches are no longer sufficient. TALM~\cite{shen2025talm} introduces a new paradigm, moving away from conventional linear workflows and towards a dynamic tree-based architecture. By using divide and conquer approaches and taking advantage of the agent's long-term memory, TALM demonstrates the need for a hierarchical structure, which can handle the non-linear dependencies involved in code generation.

In addition to this orchestration of tasks, agents should navigate the complex information space of software, which resembles a knowledge graph of dependencies and logic. While Reflexion~\cite{shinn2023reflexion} introduced a form of basic self-correction through a verbal feedback loop, recent research has focused on structural context. In this regard, RepoAudit~\cite{guo2025repoaudit} attempts to solve issues related to the repository. An auditing agent is introduced that explores the codebase on its own. This is akin to a graph traversal algorithm on file dependencies. As a result, an accurate analysis is guaranteed. MemGovern~\cite{wang2026memgovern} and Multi-Agent RL Debugging~\cite{krishnamoorthy2025multi} are extensions that arrange the external information gathered from the GitHub website and the feedback loop in the form of a retrievable knowledge base. This is a clear indication of the evolution of memory associated with software agents, which is moving away from simple memory to more structured memory that allows for basic reasoning about the topology.

\subsection{Recommender Systems}
Agent memory is applied in recommender systems to address the issue of long and dynamic user histories that are difficult to handle using recommendation agents~\cite{cai2025agentic}. In practice, recommendation agents often truncate histories, letting short-term noise, e.g., accidental interactions, override stable long-term preferences. Meanwhile, fine-tuning agent parameters to track drifting user preferences is costly~\cite{liu2025recoworld}. External memory is a more efficient solution that alleviates the problem through the reduction of retraining and token costs.

The current agent-based recommender systems have adopted a three-level memory maintenance approach that starts with coarse-grained retrieval and ends with fine-grained reasoning. In the first level of history as memory, both MAP~\cite{chen2025memory} and AgentCF++\cite{liu2025agentcf++} focus on scalability. For the latter, a dual-layer approach is proposed for noise filtering and social contextualization. For the second level of structured key-value memory, interactions are structured into semantic structures. MemoCRS\cite{xi2024memocrs} and Agent4Rec~\cite{zhang2024generative} use entity-key pairs for rating association. Finally, in the third level of memory as cognition, memory is made active by reflection and planning. At this level, CRAVE~\cite{zhu2025llm} and AgentCF~\cite{zhang2024agentcf} summarize preference rules. RecMind~\cite{wang2024recmind} applies strategic multi-step planning. In order to make these dynamic systems concrete and grounded in reality, KGLA~\cite{guo2024knowledge} applies a Knowledge Graph for concrete metadata injection. MR.Rec~\cite{huang2025mr} uses RL for adaptive memory retrieval and logical reasoning.

\subsection{Financial Agents}
Financial markets are considered to be challenging for agent memory systems due to the information decay pattern, heterogeneous data streams from various sources, and the need to balance the patterns with the latest market conditions~\cite{li2025investorbench}. 
Financial agents need memory systems with high priority for recency and the ability to preserve patterns. Moreover, financial decision-making also requires interpretability and risk management, where memory systems are essential for counterfactual reasoning, optimization, and adaptation.

The first attempts have been made in individual cognitive simulation, in which FinMem~\cite{yu2025finmem} tackles cognitive bottlenecks through a layered memory structure inspired by human traders' behavior. To address the natural bias in individual AI agents, TradingGPT~\cite{li2023tradinggpt} has proposed an extension to collaborative cognitive simulation, using inter-agent debate as a mechanism for de-biasing through collective intelligence. Progressing toward professional capabilities, FinCon~\cite{yu2024fincon} has set up a managerial structure for analysts, with agents empowered with higher memory capabilities to maintain a history of actions, profit-loss sequences, and changing investment beliefs for risk management strategies. Finally, FinAgent~\cite{zhang2024multimodal} bridges the information gap by extending memory to multimodal perception, enabling agents to process K-line charts and tool-augmented data for holistic market analysis.


Current implementations of memory-augmented financial agents remain relatively limited.
Future graph-based memory holds transformative potential in finance, e.g., hierarchical graphs can enhance multi-asset portfolio management by modeling cross-asset correlations, and temporal graphs can improve risk management and tail risk analysis.

\subsection{Game Agents}
The game environment poses a major challenge for agents’ memory systems due to its dynamics, complex rules, and long-term goals—especially in the open world that requires multi-step reasoning and exploratory learning. Memory must store experiential and world knowledge (including successes and failures) and allow efficient real-time access to support skill acquisition.
Early research on game agents focused on mastering open-ended environments like Minecraft through progressively richer memory and perception mechanisms. Ghost in the Minecraft (GITM)~\cite{zhu2023ghost} employed dictionary-based memory to capture spatial layouts and crafting knowledge via text interactions. Voyager~\cite{wang2024voyager} advanced this paradigm by introducing lifelong learning, using an ever-growing library of executable code as procedural memory for skill acquisition and reuse. Jarvis-1~\cite{wang2024jarvis} further incorporated multimodal memory to align textual reasoning with visual perception, enabling mastery of over 200 tasks in Minecraft. Most recently, Optimus-1~\cite{li2024optimus} addressed long-horizon planning by organizing experience into a hierarchical directed knowledge graph coupled with an abstracted experience pool.

Recently, the paradigm has shifted towards generalist agents capable of operating across diverse environments via unified interfaces. Cradle~\cite{tan2025cradle} broke the boundaries of game-specific APIs by establishing a unified human-like interface by grounding interaction in screenshots and keyboard-mouse actions. This enables agents to play multiple commercial games and requires episodic memory to retain cross-interaction context and accumulated gameplay experience. Following this direction, SIMA~\cite{raad2024scaling} and SIMA 2~\cite{bolton2025sima} focus on instructable agents capable of executing arbitrary natural-language commands across many 3D environments. The latest version further extends this paradigm with higher-level reasoning and self-improvement, moving from passive instruction following to active skill acquisition. These advances place increasing demands on memory systems, including long-term episodic memory and mechanisms for accumulating and reusing strategies across environments. Future research should focus on expressive memory architectures such as dynamic and temporal graph construction for latent skill hierarchy discovery and time-aware causal reasoning.

\subsection{Robotics and Embodied Agents}
Embodied agents must continuously ground their decision-making in the physical world, which is dynamic and partially observable, and perform long-horizon manipulation tasks that span multiple interaction episodes. Therefore, agents embodied in physical or virtual environments have special challenges that require advanced memory mechanisms.
HELPER~\cite{sarch2023open} and MAP-VLA~\cite{li2025map} both leverage memory to bridge high-level language instructions and executable actions, but they have different ways, such as key-value memory to map natural language commands directly to robot code and a reusable memory library to retrieve and adapt specific manipulation strategies. In a further step, STRAP~\cite{memmel2025strap} improves trajectory retrieval by introducing a flexible memory that uses dynamic time warping to match variable-length motion sub-sequences within large and diverse datasets. In contrast, TrackVLA++~\cite{liu2025trackvla++} uses a dual memory architecture to target perception stability and address the long-term maintenance.


In conclusion, the memory systems used by embodied agents remain flat or weakly structured, making it difficult to perform complex relational reasoning and hierarchical abstraction. Future work could focus on graph-based memory architectures that represent spatial relations between objects, hierarchical task structures, and action–effect causality.

\subsection{Medical and Health Agents}
Healthcare agents need advanced memory systems to facilitate high-risk medical decision-making, longitudinal patient care, and evidence-based diagnosis. Memory helps the healthcare agents retain patient histories over multiple visits, integrate new knowledge in the medical field, perform differential diagnosis, and provide personalized patient care while ensuring safety and interpretability. Specifically, AgentClinic~\cite{schmidgall2024agentclinic} and AgentMental~\cite{hu2025agentmental} focus on simulating physician–patient interactions with a memory module to record and track diagnostic information over time, while AgentMental~\cite{hu2025agentmental} employs a dynamic tree-structured memory to organize and manage medical knowledge and conversation data. The memory in healthcare agents reflects the importance of longitudinal medical histories and diagnostic accuracy in real clinical settings, facilitating agents to keep previous information to give comprehensive suggestions. In contrast, AgentHospital~\cite{li2024agent} emphasizes a full-process virtual hospital environment, where agents evolve through large-scale interactions across both successful and failed cases, supported by richer data integration and more comprehensive system functionalities.


Collectively, these studies underscore the importance of structured and interaction-aware memory for healthcare agents. Graph-based memory grounded in medical ontologies such as UMLS~\cite{bodenreider2004unified} enables multi-hop clinical reasoning, explicit modeling of drug interactions, and temporal tracking of disease progression and treatments. Such structured representations provide a promising foundation for more robust medical agents with improved contextual reasoning, diagnostic planning, and long-term decision support~\cite{moritz2025coordinated}. 


\subsection{Science Agents}
In scientific discovery, agents are increasingly shifting from passive data analysis to active experimental partners that can efficiently search vast spaces and call some specific tools. Agent memory enables the iterative integration of theory and experiment within complex scientific workflows, functioning as a dynamic workspace rather than static information retrieval. ChatNT~\cite{de2025multimodal}, ChemCrow~\cite{m2024augmenting}, and El Agente~\cite{zou2025agente} represent domain-specific scientific research support agents. All focus on enhancing LLM capabilities for professional research tasks while they have different scopes for research works such as biological sequence reasoning, chemical analysis, and quantum chemistry simulations. These systems primarily function as intelligent tools that assist researchers in localized stages of the scientific process rather than modeling the entire research lifecycle.
Furthermore, some complex agents are established as system-level scientific agents with complex environments. In biological research domain, Biomni~\cite{huang2025biomni} advances beyond task-level assistance by supporting complex biomedical research workflows, enabling agents to automatically identify, compose, and execute multi-step experimental pipelines. Similaly, CRESt~\cite{zhang2025multimodal} targets materials science field and closes the loop between computational reasoning and physical experimentation, enabling agents to iteratively generate hypotheses and validate them through robotic synthesis. Generally, VirtualLab~\cite{swanson2025virtual} operates at the organizational level, simulating human research institutions through teams of collaborative agents, and represents a shift from individual scientific tools to collective scientific intelligence, as a comprehensive agent.


Overall, this progression of existing scientific agents reflects a transition from localized scientific assistance toward holistic and autonomous scientific systems and requires effective memory modules. As agent capabilities scale, memory requirements become correspondingly more complex, demanding structured integration, organization, and management of heterogeneous domain knowledge, for which graph-based memory structures provide a clear and effective solution.

\section{Limitations and Future Directions} \label{sec:limitations}

Despite significant progress in graph-based agent memory, several fundamental challenges present critical opportunities for advancing the field.

\textbf{The Quality of Memory Graph.}
The quality of the memory graph fundamentally constrains the performance, reliability, and adaptability of graph-based agent memory systems~\cite{xiong2025memory,bei2025graphs}. Unlike traditional memory systems, where quality is often related primarily to factual accuracy in downstream tasks, graph-based memory should introduce multidimensional quality criteria, including structural, semantic, temporal, and operational aspects, each of which directly shapes agent capabilities~\cite{rasmussen2025zep,tan2025membench}. And there is a scarcity of metrics designed to explicitly evaluate the intrinsic quality of the memory graph~\cite{tan2025membench,du2025rethinking}.


\textbf{Scalability and Efficiency.} 
As agents accumulate experiences over extended interactions, memory systems face computational bottlenecks, with graph operations exhibiting quadratic or worse complexity~\cite{zeng2024structural}. Future research should explore memory compression techniques specifically designed for graph structures~\cite{leskovec2009community}, incremental update algorithms that avoid full recomputation~\cite{fan2017incremental}, and approximate retrieval methods that trade precision for substantial efficiency gains~\cite{malkov2018efficient}. Hardware acceleration through specialized graph processing units~\cite{ahn2015scalable} and distributed  architectures~\cite{shao2024distributed} enable the management of millions of nodes while maintaining rapid access.

\textbf{Privacy Protection and Security.} 
Personal assistant applications require robust protection of sensitive information while enabling meaningful personalization~\cite{li2024personal}. Graph-based memory structures introduce unique vulnerabilities where relational patterns may inadvertently expose private data through inference attacks. Critical research directions include: (1) developing differential privacy mechanisms~\cite{mueller2022sok} tailored for graph memory systems; (2) federated architectures enabling on-device processing to minimize data exposure; and (3) secure multi-party computation protocols that allow agents to benefit from collective experiences without compromising individual privacy. Beyond privacy leakage, memory systems face emerging threats from adversarial attacks. Similar to prompt injection and data poisoning attacks against LLMs~\cite{zou2023universal,carlini2024poisoning}, adversaries can manipulate memory contents to corrupt agent behavior or inject malicious knowledge. Defense mechanisms such as memory content validation, anomaly detection, and robust auditing protocols are essential to ensure memory integrity.

\textbf{Dynamic Schema Learning and Knowledge Transfer.} 
Current graph schemas are often domain-specific with limited reusability, requiring substantial re-engineering for new applications~\cite{shang2025agentsquare}. Future systems should pursue dynamic schema learning, where agents automatically identify relevant entity types and relationship patterns from raw experiences. Meta-learning approaches could enable rapid adaptation to new domains~\cite{finn2017model}, combining with universal graph ontologies~\cite{hogan2021knowledge} and domain-agnostic abstraction mechanisms, to facilitate effective knowledge transfer across tasks.


\textbf{Interpretability and Trustworthy.} 
For agents to be deployed in high-stakes domains, their memory systems must be both human-understandable and transparent in operation. Graph-based memory architectures offer unique advantages for interpretability: their explicit relational structures naturally align with human mental models, enabling users to inspect and comprehend how agents organize and utilize information~\cite{luo2024reasoning,sun2024thinkongraph,bei2025graphs}. Critical directions include developing memory provenance tracking systems, creating interactive visualization interfaces that allow users to explore memory graphs at multiple levels~\cite{huang2023memory}. By ensuring human oversight and understanding of agent memory processes, the systems can foster appropriate trust calibration, enabling users to identify potential biases, verify critical information, and maintain control over agent behavior~\cite{battaglia2018relational,hu2025agentmental}.

\textbf{Theoretical Foundations.}
Establishing rigorous mathematical frameworks remains essential for advancing the field. Priority areas include formal models that provide completeness and consistency guarantees, complexity analysis establishing theoretical bounds on construction and retrieval operations, and scaling laws of the memory-augmented AI agents~\cite{kaplan2020scaling}. Comparative analysis with human cognitive architectures could identify fundamental gaps and opportunities for architectural improvements aligned with biological memory systems~\cite{hassabis2017neuroscience}.

\textbf{Memory Coordination in Multi-Agent Systems.}
In multi-agent or agent-swarm settings, memory is no longer an isolated component but a shared resource that directly affects task completion and coordination efficiency. Ineffective memory sharing or inconsistent memory updates can lead to conflicting decisions. Designing mechanisms for memory synchronization, role-aware memory access, and scalable coordination remains an open challenge, especially under communication constraints.

\section{Conclusion} \label{sec:conclusion}
As LLM-based agents evolve toward increasingly autonomous and general-purpose systems, memory emerges as a critical component.
Graph-based memory architectures represent a paradigm shift from simple storage mechanisms to structured, relational representations that enable sophisticated reasoning, personalization, and continual learning.
This survey comprehensively reviews agent memory from a graph-based perspective. First, it introduces an agent memory taxonomy, including short-term vs. long-term memory, knowledge vs. experience memory, non-structural vs. structural memory, with a focus on graph-based memory implementation. Second, it systematically analyzes key graph-based agent memory techniques by lifecycle, including extraction, storage, retrieval, and evolution. Third, it summarizes open-source libraries, datasets and benchmarks, and diverse application scenarios for self-evolving agent memory. Finally, it identifies challenges and future research directions.
We hope this survey serves as a valuable resource for researchers advancing the frontiers of agent memory systems and for practitioners seeking to build more capable, reliable, and trustworthy AI agents.

\bibliographystyle{IEEEtran}

\bibliography{reference}

\clearpage
\clearpage
\appendices

\section{Preliminaries}

This section provides foundational concepts and formal definitions necessary for understanding graph-based agent memory systems. Since graphs serve as the fundamental structure, we begin with graph theory fundamentals, then introduce LLM-based agent architectures, and finally formalize memory components. 
Before we give the formal definitions, the frequent symbols are listed in Table~\ref{tab:definitions}

\begin{table*}[h]
\centering
\caption{Notations for Graph Foundations}
\label{tab:definitions}
\begin{tabular}{lll}
\toprule
\textbf{Symbol} & \textbf{Formal Definition} & \textbf{Meaning} \\
\midrule
$G$ & $G = (V, E, X)$ & Graph represented by node set, edge set, and node features \\
$V$ & $V = \{v_1, \dots, v_{|V|}\}$ & Set of nodes (vertices) \\
$E$ & $E \subseteq V \times V$ & Set of edges encoding pairwise relations between nodes \\
$X$ & $X \in \mathcal{X}$ & Node feature set, consisting of vectors or unstructured texts \\
$v_i, v_j$ & $v_i, v_j \in V$ & The $i$-th and $j$-th nodes \\
$e_{ij}$ & $e_{ij} = (v_i, v_j)$ & Edge from node $v_i$ to node $v_j$ \\
$\mathcal{N}(v_i)$ & $\mathcal{N}(v_i) = \{v_j \mid e_{ij} \in E\}$ & Neighborhood set of node $v_i$ \\
$d(v_i)$ & $d(v_i) = |\mathcal{N}(v_i)|$ & Degree of node $v_i$ \\
$A$ & $A \in \mathbb{R}^{|V| \times |V|}$ & Adjacency matrix representing graph structure \\
$A_{ij}$ & $A_{ij} \in \{0,1\} \ \text{or} \ \mathbb{R}$ & Adjacency matrix entry encoding the relation between $v_i$ and $v_j$ \\
$w_{ij}$ & $w_{ij} \in \mathbb{R}$ & Weight associated with edge $e_{ij}$ (if applicable) \\
$\mathbf{h}_i$	& $\mathbf{h}_i \in \mathbb{R}^d$	& $d$-dimensional embedding of node $v_i$ \\
\bottomrule
\end{tabular}
\end{table*}

\subsection{Graph Foundations}


\paragraph{Graph Definition}
We define a graph as $G = (V, E, X)$, where $V$ denotes a set of nodes, $E \subseteq V \times V$ denotes a set of edges encoding pairwise relations between nodes, and $X$ denotes node-associated features.

The graph structure is represented by an adjacency matrix $A \in \mathbb{R}^{|V| \times |V|}$, where each entry $A_{ij}$ characterizes the relation between nodes $v_i$ and $v_j$. If $A_{ij} \in \{0,1\}$, the graph is unweighted and indicates the absence or presence of an edge. If $A_{ij} \in \mathbb{R}$, the graph is weighted, and the corresponding edge weight is denoted by $w_{ij}$.

Node features $X$ may consist of continuous vectors or unstructured texts. In text-attributed graphs, $X$ corresponds to textual descriptions associated with nodes, and edge relations are typically binary and undirected, such that $e_{ij} = e_{ji}$. In knowledge graphs, node features may be represented as texts or vectors, while edges encode semantic relations and are generally directed, such that $e_{ij} \neq e_{ji}$. Knowledge graphs extensively use edge labels to represent relationship types (e.g., "located\_in," "has\_property"). When edge weights are present, they further quantify the strength or confidence of relations.

\subsubsection{Different Graphs}
\paragraph{Knowledge Graph}
A \textbf{knowledge graph} is an instantiation of the unified graph representation $G=(V,E,X)$ in which nodes correspond to entities and edges encode typed semantic relations. It is commonly formalized as $\mathcal{G}=(\mathcal{E},\mathcal{R},\mathcal{T})$, where $\mathcal{E}$ denotes entities, $\mathcal{R}$ denotes relation types, and $\mathcal{T}\subseteq\mathcal{E}\times\mathcal{R}\times\mathcal{E}$ is a set of relational triples. Each triple $(h,r,t)$ corresponds to a directed edge $e_{ij}$ from $v_i=h$ to $v_j=t$, with relation semantics attached to the edge. Knowledge graphs are typically directed and multi-relational, with node features $X$ represented as textual descriptions, learned embeddings, or both.

\paragraph{Temporal Graph}
A \textbf{temporal graph} extends the static graph $G=(V,E,X)$ by associating edges with time-dependent information drawn from a time domain $T$. Each edge $e_{ij}\in E$ is linked to a timestamp or interval $\tau(e_{ij})\subseteq T$, enabling the adjacency structure $A_{ij}$ to vary over time. This formulation supports dynamic neighborhood definitions and temporal reasoning over evolving interactions, event sequences, and state transitions, while preserving the same node feature representation $X$.

\paragraph{Hypergraph}
A \textbf{hypergraph} generalizes the pairwise edge structure of $G=(V,E,X)$ by allowing each edge to connect more than two nodes. Formally, edges are defined as subsets of vertices, i.e., $E\subseteq 2^V$, enabling the representation of higher-order and multi-way relations that cannot be reduced to binary interactions. Node features $X$ remain associated with individual vertices, while neighborhood relations are induced by shared hyperedge membership rather than pairwise adjacency.

\paragraph{Graph Variants}
Several commonly used graph variants can be viewed as task-driven instantiations of the unified representation $G=(V,E,X)$, obtained by imposing specific constraints on node and edge definitions. \textbf{Binary graphs} restrict edge values to $A_{ij}\in\{0,1\}$ and model relation existence only. \textbf{Text-attributed graphs} associate nodes with textual features, enabling language-aware representation learning. \textbf{Chunk-based graphs} further decompose text into finer-grained units as nodes, with edges encoding contextual or semantic connections. \textbf{Hierarchical graphs} introduce structural constraints to represent multi-level relationships. These variants share the same underlying graph formulation and differ only in how $V$, $E$, and $X$ are instantiated for practical modeling purposes.
These graph variants provide flexible abstractions for retrieval, representation learning, and classification. In agent graph memory systems, they support efficient organization and reasoning over heterogeneous information while maintaining a unified structural foundation.

\subsubsection{Graph Algorithms}

\paragraph{Graph Embeddings}
Graph embeddings aim to encode nodes into continuous vector spaces while preserving graph structural or semantic information. Given a graph $G=(V,E,X)$, node embeddings are defined as a mapping
\[
f: V \rightarrow \mathbb{R}^d,
\]
where each node $v_i \in V$ is associated with a $d$-dimensional representation $\mathbf{h}_i = f(v_i)$.

\paragraph{Topology-based embeddings.}
A representative traditional approach is \emph{Node2Vec}, which learns node embeddings by optimizing a Skip-gram objective over node sequences generated by biased random walks. Formally, the embedding $\mathbf{h}_i$ is learned by maximizing $\log \Pr(v_j \mid v_i)$: 
\[
\Pr(v_j \mid v_i) =
\frac{\exp(\mathbf{h}_i^\top \mathbf{h}_j)}
{\sum_{v_k \in V} \exp(\mathbf{h}_i^\top \mathbf{h}_k)},
\]
where $\mathcal{N}_l(v_i)$ denotes the context nodes of $v_i$ induced by biased random walks of length $l$. This formulation captures graph structure through node co-occurrence patterns.

\paragraph{GNN-based embeddings.}
Graph Neural Networks compute node embeddings via iterative neighborhood aggregation. At layer $k$, the embedding of node $v_i$ is updated as
\[
\mathbf{h}_i^{(k+1)} =
\sigma\!\left(
\mathbf{W}^{(k)}
\cdot
\text{AGG}^{(k)}\!\left(
\{\mathbf{h}_j^{(k)} \mid v_j \in \mathcal{N}(v_i)\}
\right)
\right),
\]
where $\text{AGG}(\cdot)$ is a permutation-invariant aggregation function, $\mathbf{W}^{(k)}$ is a learnable parameter matrix, and $\sigma(\cdot)$ denotes a non-linear activation.

\textbf{LM-based embeddings.}
For a text-attributed node $v_i$ with associated token sequence $X_i$, a Transformer-based language model encodes the input as
\[
\mathbf{H}_i = \text{Transformer}(X_i),
\]
where $\mathbf{H}_i \in \mathbb{R}^{L \times d}$ denotes the final-layer token representations. The node embedding is defined as
\[
\mathbf{h}_i = \mathbf{H}_{i,\texttt{[CLS]}},
\]
i.e., the representation of the special classification token, following standard practice in BERT-style models.


\paragraph{Graph Traversal}
Graph traversal methods define systematic or stochastic procedures for exploring the topology of a graph $G=(V,E)$, with the goal of discovering reachable nodes, structural paths, or task-relevant subgraphs that support retrieval, reasoning, and representation learning.
\begin{itemize}[leftmargin=*, topsep=0pt, partopsep=0pt, parsep=0pt, itemsep=0pt]

    \item \textbf{Breadth-first search (BFS)} explores the graph in increasing order of shortest-path distance from a source node $v_i$, iteratively visiting nodes in successive neighborhoods $N_1(v_i), N_2(v_i), \dots$, and is commonly used for local neighborhood expansion and shortest-path discovery in unweighted graphs.
    
    \item \textbf{Depth-first search (DFS)} recursively explores a path by traversing adjacent nodes as deeply as possible before backtracking, enabling efficient enumeration of paths, connectivity analysis, and cycle detection.
    
    \item \textbf{Random walk} defines a stochastic traversal process where a node sequence $(v_0, v_1, \dots, v_k)$ is generated according to transition probabilities
    \[
    p(v_{j+1}=v \mid v_j) = \frac{A_{v_j v}}{\sum_{u \in N(v_j)} A_{v_j u}},
    \]
    and serves as the theoretical foundation for classical node embedding methods.
    
    \item \textbf{Shortest-path traversal} seeks a path $P=\langle v_0=s,\dots,v_k=t\rangle$ that minimizes the accumulated edge weight
    \[
    \sum_{i=0}^{k-1} w_{v_i v_{i+1}},
    \]
    where $w_{ij}$ denotes the weight associated with edge $e_{ij}$.
    
    \item \textbf{Subgraph extraction} identifies a local induced subgraph centered at $v_i$, typically defined as the $k$-hop neighborhood
    \[
    N_k(v_i)=\{v_j \in V \mid d(v_i,v_j)\le k\},
    \]
    which is widely used for localized retrieval and downstream graph-based modeling.
\end{itemize}

\subsection{LLM-based Agents}

Table~\ref{tab:agent-notation} summarizes the key symbols used in the agent system and graph-based memory formulation, together with their semantic interpretations.

\begin{table}[h]
\centering
\small
\begin{tabular}{ll}
\hline
\textbf{Symbol} & \textbf{Description} \\
\hline
$\mathcal{A}$ & LLM-based agent \\
$\mathcal{S}$ & Environment state space \\
$s_t$ & Environment state at time step $t$ \\
$Q$ & Task specification \\
$t$ & Discrete interaction time step \\
$a_t$ & Action selected by the agent at time $t$ \\
$o_t$ & Partial observation received by the agent at time $t$ \\
$r_t$ & Feedback or reward signal at time $t$ \\
$h_t$ & Agent interaction history up to time $t$ \\
$\Psi$ & Environment transition process \\
$O(\cdot)$ & Observation function \\
$\mathcal{M}_\theta$ & Parameterized LLM policy with parameters $\theta$ \\
$\mathcal{M}$ & Agent memory module \\
$c_t$ & Retrieved memory context at time $t$ \\
$q_t$ & Memory query derived from task or observation \\
$\mathcal{M}_G$ & Graph-based memory module \\
$G_t$ & Memory graph at time $t$ \\
$V_t$ & Set of memory nodes at time $t$ \\
$E_t$ & Set of edges (relations) at time $t$ \\
$X_t$ & Node and edge attributes of the memory graph \\
$v_i$ & A memory node \\
$e_{ij}$ & A directed or undirected relation between $v_i$ and $v_j$ \\
$\mathcal{D}$ & Static corpus/experience set for memory construction \\
$\Delta_t$ & Online memory update signal $(o_t, a_t, r_t)$ \\
\hline
\end{tabular}
\caption{Notation used in the agent interaction loop and graph-based memory framework.}
\label{tab:agent-notation}
\end{table}

\subsubsection{Agent System}
An \textbf{LLM-based agent} is a decision-making system that employs a large language model as its central reasoning component to interact with an environment and accomplish a given task. Formally, an agent $\mathcal{A}$ operates over an environment with state space $\mathcal{S}$ and task specification $Q$.

At each time step $t$, the agent selects an action $a_t$ that drives the evolution of the environment via a transition process $\Psi$, based on partial observations $o_t$ rather than full access to the underlying state. The agent’s behavior is governed by a policy induced by a parameterized language model, which integrates observations, interaction history $h_t$, and optionally external memory or tools to support reasoning and decision-making, while the concrete operational procedures are specified by the agent interaction loop introduced subsequently. 

\paragraph{Agent Interaction Loop}

Specifically, during the time step $t$, the agent interacts with the environment through a structured perception–retrieval–reasoning–action-update loop:

\begin{itemize}[leftmargin=*, itemsep=0pt]
    \item \textbf{Perception.} The agent receives an observation
    \[
    o_t = O(s_t, h_t, Q),
    \]
    which encodes partial information about the environment state $s_t$ under task specification $Q$.

    \item \textbf{Retrieval.} Given the current observation and interaction history, the agent queries its external memory:
    \[
    c_t = \text{Retrieve}(\mathcal{M}, o_t, h_t),
    \]
    where $c_t$ denotes the retrieved contextual information.

    \item \textbf{Reasoning.} The LLM backbone integrates the observation, retrieved memory, and history to perform reasoning and decision-making:
    \[
    a_t \sim \mathcal{M}_\theta(o_t, c_t, h_t).
    \]

    \item \textbf{Action.} The selected action $a_t$ is executed in the environment, inducing a state transition
    \[
    s_{t+1} \sim \Psi(s_{t+1} \mid s_t, a_t).
    \]

    \item \textbf{Update.} The agent incorporates new experience and feedback into memory:
    \[
    \mathcal{M} \leftarrow \text{Update}(\mathcal{M}, o_t, a_t, r_t).
    \]
\end{itemize}

The memory module $\mathcal{M}$ stores a collection of experience tuples and knowledge representations, and supports retrieval and update operations. Its internal structure is left unspecified at this stage and may be instantiated as a key-value store, episodic buffer, or structured graph memory in later sections.

\subsubsection{Prompt Construction from Memory}

In LLM-based agent systems, memory influences agent behavior primarily through prompt conditioning. At each time step $t$, the agent constructs a composite prompt that integrates system-level instructions, retrieved memory content, and the current task context. Formally, the prompt can be abstracted as
\[
\text{Prompt}_t = \text{System}(I) \oplus \text{Memory}(c_t) \oplus \text{Task}(o_t),
\]
where $I$ specifies the agent’s role and operational constraints, $c_t$ denotes the retrieved memory context, and $o_t$ represents the current observation or query.

The memory context $c_t$ is obtained by querying the memory module with a task-dependent query $q_t$, typically through similarity-based retrieval:
\[
c_t = \text{TopK}_{m_i \in \mathcal{M}}\big(\text{sim}(q_t, m_i)\big),
\]
where $m_i$ denotes individual memory units and $\text{sim}(\cdot,\cdot)$ is a similarity function defined over their representations. This formulation provides a unified interface through which stored experiences and knowledge are incorporated into the agent’s reasoning process.


\subsection{Graph-based Memory}
In LLM-based agent systems, memory serves as a persistent mechanism for storing, organizing, and retrieving past experiences and knowledge. When memory is instantiated in a structured form, it can be naturally represented as a graph, which enables relational modeling, efficient retrieval, and incremental updates over time.

\subsubsection{Memory as Graph Format}

We formalize graph-based agent memory as a dynamic attributed graph
\[
\mathcal{M}_G \triangleq G_t = (V_t, E_t, X_t),
\]
where $V_t$ denotes the set of memory nodes, $E_t \subseteq V_t \times V_t$ denotes the set of relations between nodes, and $X_t$ represents node- and edge-associated attributes. This formulation is consistent with the basic graph foundation introduced earlier, while allowing task-specific instantiations. 

In details, each node $v_i \in V_t$ corresponds to a memory unit, such as an entity, event, concept, or textual chunk. Each edge $e_{ij} \in E_t$ represents a relation between memory units, which may encode semantic, temporal, or causal dependencies. Node attributes $X_t(v_i)$ typically include textual content or vector embeddings, while edge attributes may include relation types, confidence scores, or temporal information.

\subsubsection{Graph Memory Mechanism}

\paragraph{Graph Memory Construction}
Memory construction refers to the initial formation of a graph-structured memory from unstructured or semi-structured information, independent of online agent actions. Formally, given a corpus or experience set $\mathcal{D}$, graph construction defines a mapping
\[
\text{Construct}: \mathcal{D} \rightarrow G_0 = (V_0, E_0, X_0),
\]
where nodes $V_0$ are extracted memory units and edges $E_0$ encode relations among them.

Construction typically involves (i) node extraction, where entities, events, concepts, or textual chunks are identified as nodes, and (ii) relation extraction, where semantic, temporal, or structural relations are identified to form edges. The resulting graph may be represented as triples $(v_i, e_{ij}, v_j)$ and serves as the initial memory state.

\paragraph{Graph Memory Retrieval}
Memory retrieval corresponds to querying the graph to identify a relevant subset of nodes or subgraphs:
\[
\text{Retrieve}: (q, G_t) \rightarrow \mathcal{S}_t \subseteq V_t,
\]
where the query $q$ may be textual, structural, or embedding-based. Retrieval can be realized via semantic similarity over node representations, graph traversal from query-relevant nodes, or their combination.

\paragraph{Graph Memory Update}

Memory update describes the online evolution of an existing memory graph driven by the agent’s interaction with the environment. At time step $t$, the update signal
\[
\Delta_t \triangleq (o_t, a_t, r_t)
\]
is derived from the agent’s observation, action, and feedback. Given the current memory graph $G_t$, the update operation is defined as
\[
\text{Update}: (G_t, \Delta_t) \rightarrow G_{t+1}.
\]

The update process integrates newly observed information into the graph by adding or modifying nodes and edges, adjusting their attributes, or revising relations to reflect newly acquired evidence. Unlike memory construction, which forms an initial graph from static data, memory update operates incrementally and enables continual adaptation of the memory structure during agent interaction.



\subsubsection{Graph Memory Quality}

To systematically evaluate graph-based memory in agent systems, it is necessary to assess both the quality of memory retrieval and the structural properties induced by the graph formulation, as well as their impact on downstream task performance. We summarize representative evaluation criteria along three complementary dimensions.

\paragraph{Retrieval Effectiveness}
Retrieval quality measures the ability of the memory graph to surface relevant information in response to a query. Common metrics include Precision@K and Recall@K, which quantify relevance among the top-ranked retrieved nodes, and Mean Reciprocal Rank (MRR), which captures the ranking position of the first relevant memory.

\paragraph{Graph Structural Quality}
Graph quality metrics evaluate whether the constructed memory graph provides a coherent and faithful representation of stored knowledge. Typical criteria include coherence, which reflects structural consistency and connectivity; completeness, which measures coverage of salient information; redundancy, which captures unnecessary duplication; and temporal consistency, which assesses whether temporal relations are correctly preserved.

\paragraph{Task-level Utility}
Task performance metrics evaluate the functional usefulness of graph memory in agent decision-making, including task success rate, interaction efficiency, and generalization to unseen tasks or domains.

\section{Open-sourced Libraries}\label{app:open-sourced libraries}

In Table~\ref{tab:library_checklist}, we provide a systematic comparison of eleven representative open-source memory libraries across key functional dimensions. The columns capture essential aspects of memory systems, including license type, construction mode (external knowledge-based or interaction-driven), support for graph-based memory, retrieval mechanisms, lifecycle management, temporal modeling, reasoning capabilities, conditioning, personalization, hierarchical structure, and integration with agent frameworks. This structured overview allows for a nuanced comparison of memory design choices in agent memory design.

\begin{table*}[ht]
\centering
\caption{Comparison of Open-sourced Libraries for Graph-based Memory Systems}
\label{tab:library_checklist}
\resizebox{\textwidth}{!}{
\begin{tabular}{cp{2cm}|cc|ccccccccccc}
\toprule
\textbf{ID} & \textbf{Library} 
& \multicolumn{2}{c|}{\textbf{License}} 
& \textbf{External}
& \textbf{Interaction}
& \textbf{Graph}
& \textbf{Retrieval}
& \textbf{Lifecycle}
& \textbf{Temporality}
& \textbf{Reasoning}
& \textbf{Conditioning}
& \textbf{Personalization}
& \textbf{Hierarchy}
& \textbf{Agent}
\\
& 
& \textbf{Apache2.0} 
& \textbf{MIT} 
& \textbf{Construct.} 
& \textbf{Construct.} 
& \textbf{Memory} 
& 
& 
& 
& 
& 
& 
&  
& \textbf{Integration} \\
\midrule

1 & Cognee\footnotemark[3]    
& \ding{51} &  
& \ding{51} & 
& \ding{51} & \ding{51} 
&  &  &  &  &  &  &   \ding{51} \\
\midrule

2 & LangMem\footnotemark[4]   
&  & \ding{51}
&  &  \ding{51}
&  & \ding{51} 
& \ding{51}  &  &  & \ding{51} 
& \ding{51}  &  &   \ding{51} \\
\midrule

3 & Mem0\footnotemark[5]     
& \ding{51} &  
&  & \ding{51}
& \ding{51} & \ding{51} 
& \ding{51} 
&  &  & \ding{51} 
& \ding{51} 
& \ding{51} 
&   \ding{51} \\
\midrule

4 & LightMem\footnotemark[6]  
&  & \ding{51}
&  & \ding{51}
&  &  \ding{51}
& \ding{51} &
& \ding{51}  & 
&\ding{51}   &\ding{51}
&\ding{51} \\
\midrule

5 & O-Mem\footnotemark[7]     
& \ding{51} &  
&  &\ding{51}
&  &\ding{51}
&\ding{51} &
&  &\ding{51} 
& \ding{51} & \ding{51} 
& \ding{51} \\
\midrule

6 & OpenMemory\footnotemark[8]
& \ding{51} &  
& \ding{51} & \ding{51} 
& \ding{51} & \ding{51} 
& \ding{51} & \ding{51} 
&  &\ding{51} 
&  &\ding{51} 
&\ding{51} \\
\midrule

7 & Memori\footnotemark[9]   
&\ding{51} &  
&  &\ding{51}   
&  & \ding{51}
&  &  
&\ding{51}  &  
&  &  &    \\
\midrule

8 & MemMachine\footnotemark[10]
& \ding{51} &  
&  &  
&\ding{51}  & 
&  &  
&\ding{51} &  
&  &  &    \\
\midrule

9 & Memary\footnotemark[11]    
&  & \ding{51}
& \ding{51} & \ding{51} 
& \ding{51} & \ding{51} 
&  &  
&  &\ding{51} 
&\ding{51}  &  &\ding{51} \\
\midrule

10 & Graphiti\footnotemark[12]  
& \ding{51} &  
& \ding{51} & \ding{51} 
& \ding{51} & \ding{51} 
&  & 
& \ding{51} &  
&  &  &    \\
\midrule

11 & Memvid\footnotemark[13]
& \ding{51} &  
& \ding{51} & \ding{51} 
&  & \ding{51}
& \ding{51} &\ding{51} 
&  &  &  &  &   \\
\bottomrule
\end{tabular}
}
\end{table*}

\footnotetext[3]{\url{https://docs.cognee.ai/}}
\footnotetext[4]{\url{https://langchain-ai.github.io/langmem/}} 
\footnotetext[5]{\url{https://docs.mem0.ai/open-source/overview}}
\footnotetext[6]{\url{https://github.com/zjunlp/LightMem}} 
\footnotetext[7]{\url{https://github.com/OPPO-PersonalAI/O-Mem}} 
\footnotetext[8]{\url{https://github.com/CaviraOSS/OpenMemory}} 
\footnotetext[9]{\url{https://github.com/GibsonAI/Memori}} 
\footnotetext[10]{\url{https://github.com/MemMachine/MemMachine}}
\footnotetext[11]{\url{https://github.com/kingjulio8238/Memary}}
\footnotetext[12]{\url{https://github.com/getzep/graphiti}}
\footnotetext[13]{\url{https://github.com/memvid/memvid}}

\end{document}